\documentclass{article}

\usepackage[preprint]{neurips_2024}
\usepackage{graphicx}       
\usepackage{float}
\usepackage{amsmath}
\usepackage{mathtools}
\usepackage{enumitem}

\usepackage[utf8]{inputenc} 
\usepackage[T1]{fontenc}    
\usepackage{url}            
\usepackage{booktabs}       
\usepackage{amsfonts}       
\usepackage{nicefrac}       
\usepackage{microtype}      
\usepackage{xcolor}         
\usepackage{hyperref}
\hypersetup{
    colorlinks=true, 
    linkcolor=black, 
    urlcolor=blue,   
    citecolor=blue   
}
\title{Cognitive Memory in Large Language Models}

\author{%
Lianlei Shan, Shixian Luo, Zezhou Zhu, Yu Yuan, Yong Wu\thanks{Team leader} \\
  Li Auto\\
  \texttt{\{shanlianlei,luoshixian,zhuzezhou,yuanyu1,wuyong\}@lixiang.com} \\
}

\begin{document}

\maketitle
\begin{abstract}
In this paper, we provide an in-depth exploration of memory mechanisms in Large Language Models (LLMs), analyzing the different types of memory and their roles within these models. Memory is a crucial element in the workflow of an intelligent agent, closely related to knowledge and profiling, yet it possesses distinct granularity and functional attributes. Despite the capabilities of LLMs in information retrieval and summarizing interactions, they lack stable and structured long-term memory. The integration of memory in AI systems is vital for providing context-rich responses, reducing hallucinations, enhancing data processing efficiency, and driving the self-evolution of AI systems.
The paper begins by introducing the \textbf{cognitive architecture of memory}, categorizing it into sensory memory, short-term memory, and long-term memory. Sensory memory in humans captures fleeting information through the senses, while in LLMs, it corresponds to input requests or prompts. Short-term memory in humans holds information briefly, whereas, in LLMs, it processes inputs within the immediate context window. Long-term memory in humans includes explicit memory (episodic and semantic) and implicit memory (procedural), while LLMs implement long-term memory through external databases, vector stores, or graph structures.
\textbf{The text-based memory} section discusses memory acquisition, management, and utilization. Memory acquisition involves selecting and compressing historical information, using methods such as memory selection and summarization. Memory management covers updating, accessing, and storing memories, as well as resolving conflicting memories. Memory utilization focuses on retrieval methods, including full-text search, SQL queries, and semantic search.
\textbf{The KV cache-based memory} section introduces various KV selection and compression strategies. KV selection methods include regularity-based summarization, score-based approaches, and special token embeddings. KV compression techniques involve low-rank compression, KV merging, and multimodal compression. Memory management strategies include offloading memory, integrating with the operating system, and using shared attention mechanisms.
The paper also explores \textbf{parameter-based memory methods}, such as LoRA, Test-Time Training (TTT), and Mixture of Experts (MoE). These methods transform memories into model parameters, enhancing memory efficiency and utilization.
Finally, the paper discusses \textbf{hidden-state-based memory}, including chunk mechanisms, recurrent transformers, and the Mamba model. These approaches combine the concept of hidden states in RNNs with current methods to improve the processing and memory efficiency of long texts.
Overall, this paper provides a comprehensive analysis of memory mechanisms in LLMs, highlighting their significance and offering guidance for future research directions.
\end{abstract}

\newpage
\tableofcontents
\newpage

\section{Introduction}
Memory, more precisely memories, constitutes a crucial foundational element of an agentic workflow and is intricately interwoven with knowledge and profiling. However, it deserves separate and focused attention due to its distinct granularity and functional attributes compared to 'knowledge' and 'profile'.
Profiling delineates how an agent comprehends its own identity (encompassing its character and 'avatar'), its functions (behavioral models), and its operational context (environment). Knowledge, on the other hand, supplies the factual information or learned representations that underpin decision-making processes. Memory, distinct from both, serves as the dynamic repository of experiences that integrates these elements and actively engages in decision-making.
Despite decades of research, the consistent memory retention capabilities of large language models (LLMs) remain an area of ongoing exploration. Contemporary AI systems are capable of information retrieval, summarization of past interactions, and selective detail storage. However, they lack a stable and structured memory that can reliably endure over extended periods.

In the realm of artificial intelligence, the integration of memory is becoming increasingly significant. As the complexity of AI systems escalates, the incorporation of memory functions confers numerous substantial advantages upon these systems, propelling advancements across multiple dimensions \cite{huggingface_memory_blog}.

Firstly, memory integration enables AI systems to provide context-rich responses. Traditional large language models (LLMs) are typically stateless, processing each prompt in isolation without retaining prior contextual information. However, LLMs augmented with memory systems break through this limitation by integrating memory, achieving continuity across interactions. They can leverage both short-term context and long-term data simultaneously, thereby delivering more in-depth and personalized responses to users, significantly enhancing user experience.
Secondly, memory integration helps to mitigate hallucination phenomena. When LLMs attempt to fill gaps in retrieval but lack relevant knowledge, hallucinations are likely to occur. By anchoring responses in stored facts, namely, through the Retrieval-Augmented Generation (RAG) technique, memory systems can effectively reduce such errors. Although the implementation of RAG poses challenges such as advanced data processing requirements, its limitations are continually being addressed. For instance, Graph-RAG methods, which utilize graph-based structures to improve retrieval accuracy and scalability, offer new insights and pathways to resolve this problem.
Thirdly, memory integration achieves efficient data processing. In practical applications, manually reviewing large volumes of documents, such as PDF files or financial statements, is not only time-consuming but also prone to errors. In contrast, LLM pipelines equipped with memory functions can automatically ingest, classify, and store data, and retrieve it on demand. This architecture supports targeted retrieval, reducing unnecessary API calls or database queries, thereby lowering computational costs, increasing work efficiency, and accelerating business processes across various industries.
Lastly, memory integration represents a crucial step for AI systems towards self-evolution. Integrating long-term memory into AI systems allows LLMs to adapt to new tasks in a manner akin to human learning, even when data or interactions are limited. This not only enhances their performance and applicability in real-world scenarios but also lays a solid foundation for the future development of AI, driving breakthroughs and innovations in more fields.

This paper will elucidate the various types of memory and their respective roles within an agentic workflow. It will also explore how these components coalesce in practical applications, clarify the mechanisms by which models in memory mode remember information, and examine the transformative impact of generative AI on the very nature of memory. We first delve into the cognitive architecture, clarifying the categorization of memory and the distinctions between the ways humans and LLMs process memories. Subsequently, we analyze the current state of research on LLM memory. Finally, we introduce the future trend of LLM memory: cognitive memory.

\section{Cognitive Architecture}
Cognitive architecture is an important tool for studying human thinking and intelligent behavior, and it has a history spanning several decades. \cite{40years} reviews the development of cognitive architectures over the past 40 years, focusing on their research and application in simulating core human cognitive abilities, such as memory, perception, attention, and decision-making. By simulating psychological experiments, such as working memory tests and attention experiments, researchers have verified the ability of cognitive architectures to model human cognitive processes.

The Soar Cognitive Architecture \cite{soar} and the ACT-R Cognitive Architecture \cite{act-r} are important works in the field of cognitive architecture. Soar is a general intelligent system that can handle various types of problems, from simple logical reasoning to complex strategic games, through its General Problem Solver (GPS) module. It also supports multiple learning mechanisms, such as rule-based learning and case-based learning. ACT-R focuses on simulating human cognitive processes, especially memory, learning, and decision-making. It studies the interaction between working memory and long-term memory through detailed memory models and can simulate human decision-making in complex tasks. These architectures have not only played an important role in artificial intelligence research but have also verified their ability to model human cognitive processes through psychological experiments. The ICARUS Cognitive Architecture \cite{icarus} represents an emerging direction in cognitive architecture research. ICARUS aims to develop intelligent systems with autonomous learning and adaptation capabilities by simulating human cognitive processes. It supports autonomous learning mechanisms, can learn new knowledge through interaction with the environment, and can handle complex tasks such as robot navigation and multitasking. The research on ICARUS has further promoted the application and development of cognitive architectures in the fields of artificial intelligence and cognitive psychology.
In cognitive architecture, memory is divided into three types: sensor memory, short-term memory, and long-term memory.

\subsection{Categories of Memories}
Human memory can be categorized from different perspectives. Firstly, based on the duration of memory retention, memory can be divided into iconic memory, short-term memory, and long-term memory. Iconic memory lasts for an extremely short period, typically between 0.25 and 2 seconds. It has a large capacity, but the information is not processed and fades quickly. For example, the fleeting impression we get when we see an object is a form of iconic memory. Short-term memory lasts for about one minute or less and has a limited capacity, usually around $7\pm2$ information units. Information in short-term memory needs to be rehearsed to be maintained, such as when we repeat a phone number to remember it. Long-term memory, on the other hand, can last from over a minute to a lifetime. It has a vast capacity and stores information that has been deeply processed, like childhood experiences and learned knowledge.

The content of memory can be classified into iconic memory, logical memory, emotional memory, and procedural memory. Iconic memory involves the images of perceived objects, such as the memory of a painting or a melody, and is characterized by its intuitiveness and concreteness. Logical memory deals with the outcomes of logical thinking, such as concepts, formulas, and laws, focusing on understanding the essence and internal connections of things, like the memory of mathematical formulas and physical laws. Emotional memory is about the emotions or feelings experienced in the past, such as happiness or sadness, and carries a deep emotional hue. Procedural memory involves the images of bodily movements or actions, such as the memory of riding a bicycle or playing the piano, and is the foundation for learning motor skills.
Based on the degree of conscious involvement, memory can be divided into explicit memory and implicit memory. Explicit memory refers to the conscious recall of past events or learned information. The content can be described in language, such as remembering places we have visited or knowledge we have acquired. Implicit memory, however, is the recall of past experiences without conscious involvement. It is usually related to motor skills and habits and does not require language description, like the unconscious memory of riding a bicycle.

In the cognitive structure, memory is divided into sensor memory, short-term memory, and long-term memory. We classify and introduce memory according to this standard.

\subsubsection{Sensor Memory}
In humans, sensory memory refers to the information we grasp through our senses (visual, auditory, or otherwise) within the first few seconds of perception, after which it’s either discarded or moved to short-term memory. In LLM systems, SM corresponds to an API request or prompt being fed into the system.
Sensory memory captures quick impressions from our surroundings, like the flash of a passing car or the sound of footsteps, but these fade almost instantly.

\subsubsection{Short-Term Memory}

\begin{figure}[H]
  \begin{center}
    \includegraphics[width=1.0\linewidth]{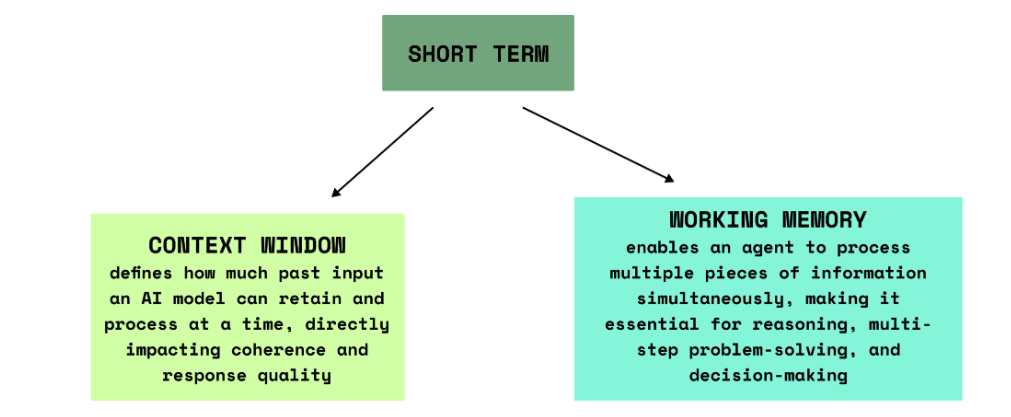}
  \end{center}
  \caption{Short Memory}
  \label{fig:short_mem}
\end{figure}

Short-term memory (STM) is defined as Figure \ref{fig:short_mem}.
Human short-term memory holds and manipulates a small amount of information in an active state. In LLMs, STM can be thought of as handling the input - the text tokens or embeddings currently available for prompt processing - within the immediate context window. Once this session ends, the model usually “loses” that context.

In the field of Natural Language Processing (NLP), the context window is a crucial concept in Large Language Models (LLMs). It refers to the range of contextual information that a model can consider when processing and generating text, typically measured in the number of tokens. For example, the context window of GPT-4o reaches 128,000 tokens, which means that it can process and generate longer and more complex texts in a single interaction.

The size of the context window has a profound impact on the model's performance. A larger context window enables the model to maintain coherence in long conversations and better track multiple threads of dialogue. For instance, in complex multi-turn dialogues, the model can remember previous content to generate more relevant and natural responses. Moreover, a larger context window helps the model handle long documents, such as summarization and question-answering tasks. It allows the model to capture long-distance dependencies and global information within the document. A larger context window also enables the model to better understand semantic relationships in the context, thus generating more accurate and coherent text.
From a technical implementation perspective, the attention mechanism is the core technology for realizing the context window. It allows the model to consider other relevant words in the text while processing a single word. The Transformer architecture and its self-attention mechanism are key to achieving a large context window. They enable the model to process all words in the text simultaneously, rather than just neighboring words. Additionally, positional encoding is used in the Transformer model to represent the position of words in the text, helping the model understand the order and relative positions of words.

Working memory plays a vital role in multi-step reasoning and decision-making. Just as humans use working memory to hold several ideas in mind at once – like when solving a math problem – AI agents rely on it to process multiple inputs simultaneously. This is especially important for complex tasks like planning, where an agent needs to balance different pieces of information before reaching a decision.

\subsubsection{Long-Term Memory}

\begin{figure}[H]
  \begin{center}
    \includegraphics[width=1.0\linewidth]{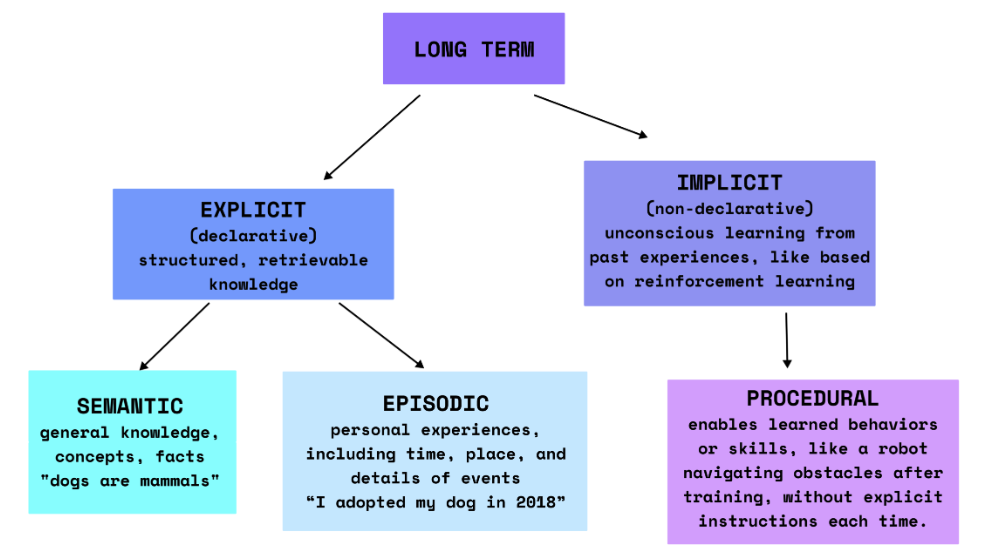}
  \end{center}
  \caption{Long Memory}
  \label{fig:long_mem}
\end{figure}

In humans, long-term memory retains knowledge, experiences, and skills over time. Long-term memory is divided into explicit (declarative) memory, which is conscious and includes episodic memory (life events) and semantic memory (facts and concepts), and implicit (procedural) memory, which is unconscious and encompasses skills and learning tasks, as shown in Figure \ref{fig:long_mem}. For LLMs, LTM can be implemented through external databases, vector stores, or graph structures that keep relevant data available and allow the model to “recall” information in future queries.

Explicit memory is what allows AI to recall facts, rules, and structured knowledge. Within this category, semantic memory is responsible for storing general truths and common knowledge. 
Semantic memory in AI mirrors its human counterpart—it involves the storage of factual information. AI systems use semantic memory to retain and recall facts and general knowledge from the training materials. For instance, a digital assistant uses semantic memory to retrieve product information from a user manual.

Episodic memory is more personal – it captures specific events and experiences, allowing an agent to remember context from past interactions. If a customer service AI recalls that a user previously requested a refund, it can tailor its responses accordingly, making interactions feel more intuitive and human-like.
Episodic memory for AI relates to the ability to contextualize past interactions or experiences. This type of memory is critical to AI models in fields like customer service, as episodic memory can recall previous customer conversations to optimize for the best solution. Furthermore, episodic memory can recall previous conversations with an individual customer to offer personalized service, much like a salesperson remembers past encounters with clients. In the service and experience fields, episodic memory allows AI to capture the tribal knowledge needed to deliver effective automation.

Implicit memory, on the other hand, is what allows AI to develop instincts, which involves the retention and recall of the steps required to complete tasks. It’s driven by procedural memory, which helps an agent learn skills without requiring explicit recall. Think of a self-driving car that improves its lane-keeping ability after thousands of miles of training. The car doesn’t need to “remember” every scenario explicitly – it develops an intuitive understanding of how to navigate roads. When you actually experience that – it’s quite incredible.
In AI, procedural memory allows systems to perform automated functions they have learned through repetitive training. One common example would be operating machinery in a factory setting. In the context of service and customer experience, procedural memory enables AI to perform multi-step sequences, effectively addressing and resolving even complex customer needs.

\subsection{Key Differences and Parallels Between Human Memory and LLMs}

\subsubsection{Human Memory}
Human memory is an intricate and essential cognitive function that is intricately intertwined with our emotional, experiential, and biological processes. It can be fundamentally categorized into three primary types: sensory memory, short-term memory, and long-term memory.

Sensory memory is responsible for capturing fleeting perceptual stimuli from the environment, such as the brief visual impression of a passing vehicle or the transient auditory cue of footsteps. However, these sensory traces typically dissipate almost immediately, lasting only a fraction of a second.

Short-term memory, in contrast, serves as a temporary storage system for information that is needed for immediate use. It allows individuals to retain small amounts of data over a brief period, typically lasting from seconds to a few minutes. An illustrative example is the act of looking up a phone number and then dialing it without needing to write it down. This process exemplifies the role of short-term memory in managing immediate tasks.

Long-term memory represents the repository of enduring knowledge, skills, and emotional experiences that define the richness of human cognition and behavior. It encompasses two main subcategories: declarative memory, which involves the retention of factual information and episodic events, and procedural memory, which pertains to learned tasks, habits, and motor skills. The transition of memories from short-term to long-term storage is known as consolidation, a process that relies heavily on the brain's neurobiological mechanisms, particularly the hippocampus. This critical brain region plays a pivotal role in strengthening and integrating memories over time, thereby facilitating their long-term retention.

Human memory is not static; it is inherently dynamic and malleable. Memories can be modified, updated, or even reconstructed based on new experiences, emotional significance, and contextual factors. This adaptability is crucial for learning and cognitive development, as it enables individuals to selectively retain relevant information while discarding less important details. However, this same flexibility also renders human memory susceptible to inaccuracies and biases. Memories are often reconstructed rather than retrieved in their exact original form, influenced by various factors such as context, emotional states, and personal interpretations. While this can occasionally lead to unreliability, it underscores the complex and adaptive nature of human memory, which fundamentally distinguishes it from the more rigid and deterministic memory systems employed in artificial intelligence.

\subsubsection{LLMs Information}
Large Language Models (LLMs) operate on fundamentally distinct principles compared to human memory when processing and storing information. These models are trained on extensive datasets that encompass diverse textual sources, including books, websites, and articles. During the training process, LLMs identify and learn the statistical patterns inherent in language, discerning the relationships between words and phrases. Unlike human memory, LLMs do not possess a memory system in the conventional sense; instead, they encode these linguistic patterns into a vast array of parameters—typically billions of numerical values. These parameters determine how the model predicts and generates responses based on input prompts.
LLMs lack explicit memory storage akin to that of humans. When presented with a query, an LLM does not retain a record of previous interactions or the specific data used during training. Rather, it generates a response by calculating the most probable sequence of words based on the patterns learned from its training data. This process is facilitated by sophisticated algorithms, particularly the transformer architecture, which incorporates an attention mechanism. This mechanism enables the model to selectively focus on relevant segments of the input text, thereby producing coherent and contextually appropriate responses.
In essence, the "memory" of LLMs is not a true memory system but rather an emergent property of their training process. They rely on the encoded patterns to generate responses, and any adaptation or learning occurs only through retraining on new data. This is a critical distinction from human memory, which evolves dynamically through continuous experiential input and is inherently adaptive in real-time.

\subsubsection{Parallels and Differences}
Despite fundamental differences in information processing between humans and large language models (LLMs), several noteworthy \textbf{parallels} exist. Both systems rely heavily on pattern recognition to process and interpret data. In humans, pattern recognition is crucial for learning, enabling the recognition of faces, comprehension of language, and recall of past experiences. Similarly, LLMs excel at pattern recognition, utilizing their training data to learn the intricacies of language, predict subsequent words in a sequence, and generate coherent text.
Context is another critical factor that influences both human memory and LLMs. For human memory, context enhances the effectiveness of information recall. For instance, being in the same environment where learning occurred can trigger related memories. LLMs also leverage the context provided by input text to guide their responses. The transformer architecture allows LLMs to selectively attend to specific tokens (words or phrases) within the input, ensuring that the response aligns with the surrounding context.

Furthermore, both humans and LLMs exhibit phenomena akin to primacy and recency effects. Humans tend to remember items at the beginning and end of a list more readily, known as the primacy and recency effects. In LLMs, this is reflected in how the model assigns greater weight to specific tokens based on their position in the input sequence. The attention mechanisms in transformers often prioritize more recent tokens, enabling LLMs to generate contextually appropriate responses, similar to how humans rely on recent information to guide recall.

However, \textbf{the differences} between human memory and LLMs are far more profound. One significant distinction lies in the nature of memory formation. Human memory is dynamic and constantly evolves, influenced by new experiences, emotions, and context. Learning new information can alter existing memories and change how they are perceived and recalled. In contrast, LLMs are static after training. Once an LLM is trained on a dataset, its knowledge remains fixed until it undergoes retraining. It cannot adapt or update its memory in real time based on new experiences.

Another key difference pertains to how information is stored and retrieved. Human memory is selective, with emotionally significant events being more likely to be remembered while trivial details fade over time. LLMs, however, lack this selectivity. They store information as patterns encoded in their parameters and retrieve it based on statistical likelihood, rather than relevance or emotional significance. This leads to a stark contrast: LLMs have no concept of importance or personal experience, whereas human memory is deeply personal and shaped by the emotional weight assigned to different experiences.
One of the most critical differences lies in the function of forgetting. Human memory features an adaptive forgetting mechanism that prevents cognitive overload and helps prioritize important information. Forgetting is essential for maintaining focus and making space for new experiences, allowing humans to discard outdated or irrelevant information and constantly update their memory.

In contrast, LLMs do not exhibit adaptive forgetting. Once trained, they retain all information within their exposed dataset. The model only updates its knowledge if it is retrained with new data. However, in practice, LLMs may appear to "forget" earlier information during extended conversations due to token length limitations. This is a technical constraint rather than a cognitive process.

Finally, human memory is closely intertwined with consciousness and intent. Humans actively recall specific memories or suppress others, often guided by emotions and personal intentions. LLMs, by contrast, lack awareness, intent, or emotions. They generate responses based on statistical probabilities without any underlying understanding or deliberate focus.

\section{Text-based Memory}

In this chapter, we introduce memory stored in the form of text. We discuss memory acquisition, memory management, and memory utilization. Memory acquisition refers to the process of selecting and compressing historical information to store it in the memory bank. This process can be categorized into three approaches: memory selection, memory summarization, and a combination of both. Memory management involves the storage, updating, and access mechanisms of the memories stored in the memory bank, as well as the resolution of conflicting memories. Memory utilization, on the other hand, pertains to memory retrieval, which is the process of obtaining the most relevant memories based on current information (such as a query). This includes full-text search, SQL-based search, semantic search, tree-based search, hash-based search, multi-pass search, and a hybrid search approach that combines multiple methods. Acquisition, management, and utilization encompass the entire process of text-based memory. The structure of this chapter is depicted in the figure above.

\subsection{Memory Acquisition}

\begin{figure}[H]
  \begin{center}
    \includegraphics[width=1.0\linewidth]{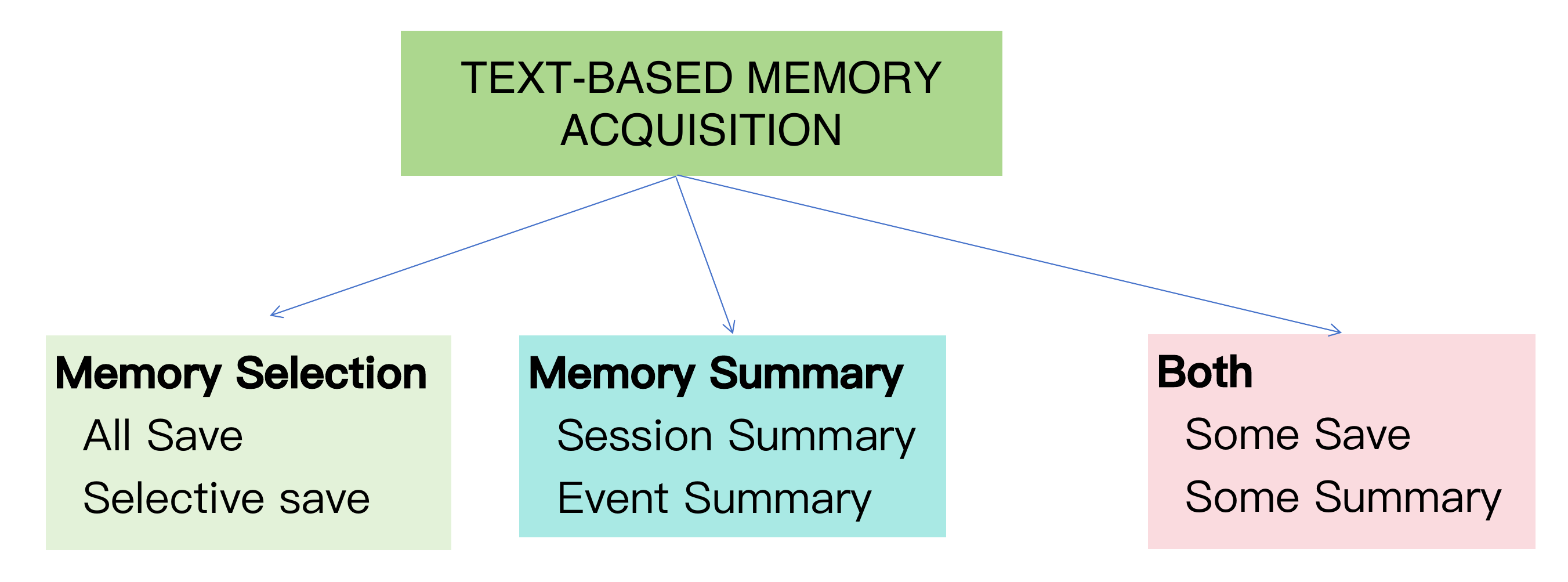}
  \end{center}
  \caption{Text-based memory acquisition.}
  \label{fig:mem_ac}
\end{figure}
Figure \ref{fig:mem_ac} shows the overall structure of memory acquisition.
Memory acquisition includes three ways: memory (or called text) selection, memory summarization, and both.

\subsubsection{Memory Selection}

Memory selection refers to the process of retaining all or selectively preserving the content of conversations or texts. The simplest approach is to retain all historical information as memory. Building on this foundation, researchers have proposed many selection methods based on predetermined strategies. For example, discard greetings or texts that lack informational content []. This is a rudimentary method and is generally used as a baseline for work. In summary, selection strategies can generally be flexibly defined according to preset criteria.

\subsubsection{Memory Summarization}

Memory summarization refers to the method of retaining the summarized content after condensing conversations or texts.

COMEDY \cite{comedy} faces the challenge of an excessively large dataset (over 500,000 data points) in extracting and generating conversation-level memories. To address this, COMEDY focuses on a subset of about 40,000 data points. GPT4-Turbo is used to extract conversation-level memories (including events, user, and robot descriptions), which are then edited by annotators to ensure the completeness and accuracy of the information, creating a collection of conversation-level memories \( M \). Additionally, COMEDY develops a dedicated LLM for generating conversation-level memories and filters out invalid samples to ensure data quality.
In the memory compression task, COMEDY uses GPT4-Turbo to summarize all conversation-level memories \( M \) from Task 1 and output the compressed memories \( \hat{M} \). This process includes creating comprehensive user profiles, capturing the dynamic interactions between users and robots, and recording key events. To balance creativity and relevance, GPT4-Turbo generates three outputs at a temperature setting of 0.9. Annotators then refine and calibrate these outputs, correcting any inaccuracies or inconsistencies, and enhancing clarity and conciseness where necessary to ensure the summaries accurately reflect user profiles, interaction dynamics, and event records.

Memory Bank \cite{memorybank} employs a hierarchical approach to event summarization and dynamic personality understanding to simulate the complexity of human memory and enhance user experience. It condenses conversations into daily event summaries and global summaries, creating a layered memory structure that provides users with a macro view of past interactions. Additionally, Memory Bank continuously updates its understanding of user personality through long-term interactions, generating daily personality insights that are aggregated into a global understanding. This method enables AI companions to tailor their responses based on individual user traits, achieving more personalized interactions.

In \cite{ahuman}, LLMs process documents by segmenting them into chunks with a maximum of max words, ensuring at least min words are processed per step. The ratio of max words to min words determines the maximum number of words an LLM can handle. Additionally, memory-based summarization extracts summaries from each chunk independently. During retrieval, parallel searching based on summaries significantly reduces processing time compared to sequential searching of raw text. The answer generation process is similar to parallel searching but incurs additional overhead due to prompt templates. Optimizing chunking and summarization strategies can effectively enhance LLM processing efficiency.

In\cite{thinkinmemory}, inductive reasoning is defined as text that includes the relationship between two entities, namely the relational triple $(E_h, r_i, E_t)$. $E_h$ is the head entity connected to the tail entity $E_t$ through the relation $r_i$, where $i \in [0, N]$ and $N$ is the number of relations. Conceptually, $R_h = \{r_1, \cdots, r_N\}$ represents all the one-hop relations of entity $E_h$. 

The main challenge in applying inductive reasoning to large language models is obtaining high-quality sentences that match the relational triple. This paper provides two solutions for acquiring inductive reasoning: (1) pre-trained models for open information extraction, such as OpenIE\cite{openie}; (2) few-shot prompt-based context learning based on large language models. In this work, the authors adopted the second solution, namely using the large language model AgentA to generate inductive reasoning.

\subsubsection{Both}


Memory construction is a key issue in memory-based models. Different forms or sources of memory can significantly affect the efficacy of the model. In \cite{evolving}, three different types of memory are explored, and their respective impacts on model performance are compared.
This study explores the impact of three types of memory on model performance. Memory based on historical records directly uses unprocessed dialogue history, which is simple and practical but prone to redundancy. Memory based on summaries provides richer context through dialogue summaries but may lose details. Conditional memory judges the importance of utterances, storing only key information and dividing memory into two parts: context and knowledge, to enhance memory efficiency and relevance. In addition, the study attempts multiperspective memory, combining various types of memory to provide more diverse information for response generation.

\subsection{Memory Management}

\begin{figure}[H]
  \begin{center}
    \includegraphics[width=1.0\linewidth]{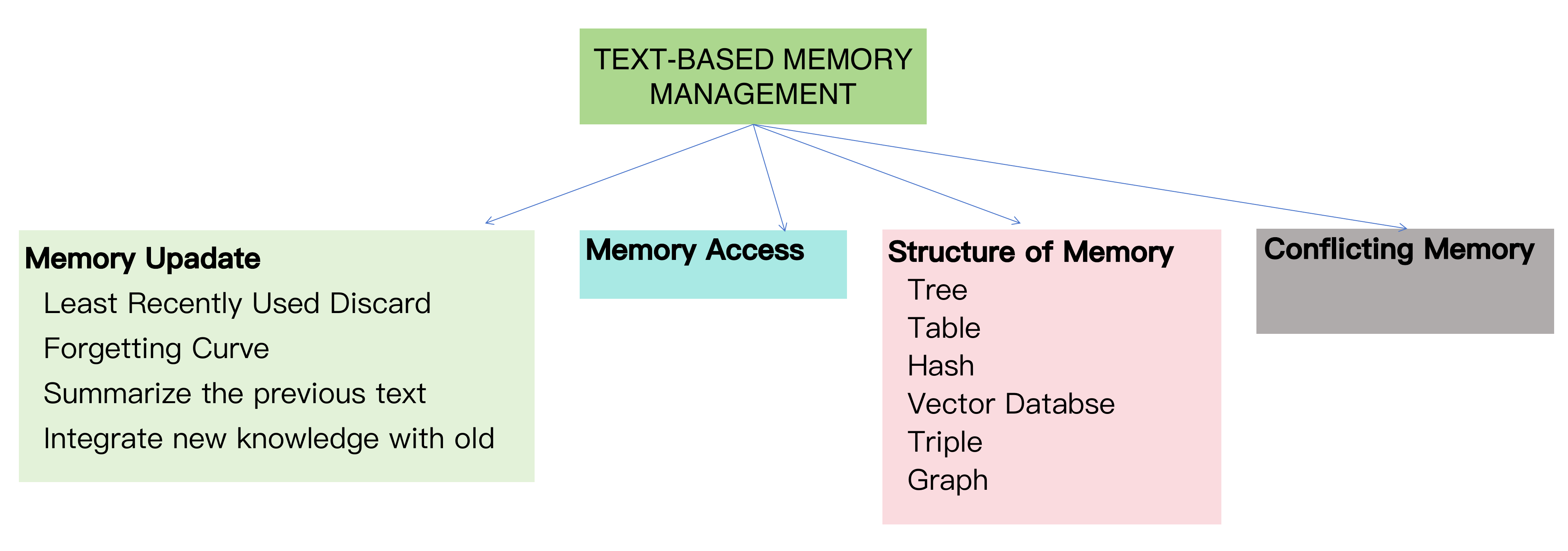}
  \end{center}
  \caption{Text-based memory management}
  \label{fig:mem_mana}
\end{figure}

Memory management encompasses four aspects: memory update, memory access, data structure for memory storage, and handling contradictory memories, as shown in Figure \ref{fig:mem_mana}.

\subsubsection{Memory Update}

The strategies for memory updating include the Least Recently Used (LRU) discard strategy, the Forgetting Curve strategy, the Previous Context Summary Update strategy, and the Integration of Old and New Knowledge strategies.

\textbf{The LRU algorithm} is a cache eviction strategy based on recent access patterns. Its core idea is to discard the data that has not been accessed for the longest time. It is typically implemented using a doubly linked list and a hash table: The head of the linked list is the most recently accessed data, while the tail is the least recently accessed data. When accessing data, if the data is in the cache, it is moved to the head of the linked list; if it is not in the cache, it is added to the head, and if the cache is full, the data at the tail of the list is evicted.


Memory forgetting mechanism is inspired by \textbf{Ebbinghaus's theory}, which shows that memory retention decreases over time, with a steeper decline initially and a slower rate later. The spacing effect suggests that relearning information is easier than learning it for the first time. Regular reviews can reset the forgetting curve and improve retention.
The forgetting curve can be modeled by an exponential decay formula:
\begin{equation}
R = e^{- t S}
\end{equation}
where \(R\) is the retention rate, \(t\) is the time since learning, \(e\) is approximately 2.71828, and \(S\) is memory strength, influenced by factors like learning depth and repetition. In \cite{memorybank} To simplify, the author models \(S\) as a discrete value, initialized at 1 and increased by 1 each time a memory is recalled, resetting \(t\) to 0. This reduces the forgetting probability, though real-life memory is more complex.
MemoryBank combines these elements to create a comprehensive memory management system for LLMs, enhancing their ability to provide meaningful, personalized interactions over time and expanding possibilities for AI applications.

\textbf{Previous Conversation Summary Update Strategy}
\cite{recursively} instructs Large Language Models (LLMs) to recursively generate memories (summaries) using dialogue context and prior memories. Specifically, the updated memory is calculated using the following formula:
\begin{equation}
M_i = \text{LLM}(H_i, M_{i-1}, P_m) 
\end{equation}
where \( M_i = \{m_1, m_2, \dots, m_S\} \) represents multiple sentences containing key information summarized from the past conversation \( H_i \), and \( P_m \) is the LLM prompt used for generating memories. Specifically, the author inputs the previous memory and the entire conversation to the LLM and asks it to update the previous memory with the prompt ``Your goal is to update the memory.'' By integrating the new information from the given dialogue context, the previous memory is \( [M_{i-1}] \), and the dialogue is \( [H_i] \). The author sets the initial memory \( M_0 \) as ``empty.'' This operation is repeated \( N \) times until the last conversation ends, at which point the final memory \( M_N \) can be obtained.

\textbf{The Strategy of Integrating New and Old Knowledge}

Inspired by human memory, it is necessary to organize dynamic updates and idea evolution based on established operating principles, such as inserting new ideas, forgetting unimportant ones, and merging repetitive ones, so as to make the memory mechanism more natural and applicable. Starting with the memory cache storage architecture, \cite{thinkinmemory} uses a hash table to store self-generated ideas, where each hash index corresponds to a group of ideas containing similar thoughts. Within the same group, Tim supports the following operations to organize the ideas in memory: Insertion, which means storing new ideas in memory; Forgetting, which means removing unnecessary ideas from memory, such as contradictory ones; and Merging, which means combining similar ideas in memory, such as those with the same head entity.

\subsubsection{Memory Access}

In \cite{empowering}, the Working Memory Hub serves as a unified hub, coordinating data flow among various components. It stores inputs, outputs, and interaction history records and provides features such as the Interaction History Window and the Episodic Buffer to address the issue of memory silos. At its core is the Central Processor based on LLMs, which is responsible for processing and analyzing information and making decisions in conjunction with inputs and outputs from the External Environment Interface. The External Environment Interface dynamically acquires real-time inputs and transmits the outputs from the Central Processor, with all data stored in the Working Memory Hub. The Interaction History Window offers short-term caching and context anchoring, while the Episodic Buffer addresses the token limitations of LLMs by enabling the retrieval of complete scenarios.

Memory access strategies include role-based access, which assigns permissions based on roles; task-based access, which provides permissions based on task requirements; and autonomous memory access, which allows agents to independently select relevant memories. In complex scenarios, the Memory Management Agent is responsible for efficiently managing, sorting, and retrieving historical data to enhance the efficiency of multi-agent systems.

\subsubsection{Data Structure for Memory Storage}
The storage structures of memory include trees, tables, hash tables, vector databases, triplets, and graphs.

\cite{enhancing} proposed a tree-based search.
The Hierarchical Aggregate Tree (HAT) is defined as $HST = (L, M, A, \Sigma)$. Here, $L$ represents the hierarchical structure of the tree, $M$ denotes the set of nodes in the tree, $A$ is the aggregation function used to aggregate text information from child nodes to parent nodes, and $\Sigma$ also represents the set of nodes in the tree. An important characteristic of the HAT is that as you move down the hierarchy, the resolution increases while moving from left to right results in more up-to-date information. This property enables the HAT to effectively manage and retrieve memory information in long text conversations. By using GPT to transform child nodes into parent nodes through prompts, the HAT can efficiently process and organize information.

\cite{toward} proposed a table-based search.
In a table-based chat database, each response contains text as well as metadata such as the speaker's name, date, time, and conversation ID. This naturally suits a tabular database representation, where each column of the table contains a specific type of data (e.g., time), and each row contains information related to a particular response. We integrate this table with a vector database by creating a “content” column, which stores the index of the relevant semantic vector in the response vector database. The table can then be queried using the top k responses found through semantic retrieval, as well as any queries based on metadata.


\cite{thinkinmemory} aims to store inductive thinking in memory according to certain rules, that is, similar thinking should be stored in the same group of memory to improve efficiency. To this end, the author adopts a hash table as the architecture of the TiM storage system, where similar thinking is assigned the same hash index. For a given query, the author proposes to quickly search for its nearest thinking in the high-dimensional embedding space, which can be solved by the Locality-Sensitive Hashing (LSH) method. The goal of the LSH hashing scheme is to assign each $d$-dimensional embedding vector $\mathbf{x} \in \mathbb{R}^d$ to a hash index $F(\mathbf{x})$, where neighboring vectors have a higher probability of obtaining the same hash index. The author achieves this by using random projection, as shown below:
\begin{equation}
    F(\mathbf{x}) = \arg \max \left([\mathbf{x} \mathbf{R}; -\mathbf{x} \mathbf{R}]\right) 
\end{equation}
where $\mathbf{R}$ is a random matrix of size $(d, b/2)$, and $b$ is the number of groups in memory. $[\mathbf{u}; \mathbf{v}]$ denotes the concatenation of two vectors.


Vector embeddings have become a popular method for representing memory because Transformer-based neural networks and large language models (LLMs) inherently use vector embeddings. Input data in its original text form is embedded into a fixed-dimensional space through an encoder network. These embeddings represent the data in a way that allows for similarity retrieval using distance-based metrics. Vector databases are designed to support the storage and retrieval of data represented as embeddings, rather than the more structured scalar data found in traditional relational databases. Queries in vector databases rely on similarity matching to return the top \( k \) matches \cite{memory}, using algorithms such as cosine similarity, Euclidean distance, or dot product. Metadata can be used as an additional filtering step to provide conditional logic filtering on top of similarity-based retrieval (e.g., the year metadata field must be greater than or equal to '2020').


In \cite{relational}, OpenIE is applied to a specific corpus to directly extract relational triples in the form of (head entity, relation, tail entity). The advantage of this approach is that it can obtain structured knowledge closely related to a specific dataset, rather than relying solely on general knowledge bases. Based on this, retrieval strategies can be designed according to different needs, such as selecting the most relevant relational triples through keyword matching, vector similarity, or other relevance measures.

Research on LLMs based on knowledge graphs explores how to integrate knowledge graphs and large language models to enhance the professionalism and accuracy of question-answering systems, as demonstrated in the application research in the field of traditional Chinese medicine. A review of the research progress in medical knowledge graph construction technology summarizes knowledge representation, extraction, fusion, reasoning, and quality assessment, and discusses its applications in medical services. A comprehensive overview of the collaborative research between large language models and knowledge graphs discusses how to combine the strengths of LLMs and knowledge graphs. Research on the automatic construction of knowledge graphs introduces methods for using COMET and GPT models to predict whether newly generated triples are reasonable. Additionally, although mainly discussing spatial data indexing and retrieval techniques, the study on fast indexing and dynamic retrieval methods for large-scale vector models based on octrees and R-trees provides insights into the efficient management and retrieval of large datasets for knowledge graph construction and management. These studies cover multiple aspects, from the automatic construction of knowledge graphs to their collaborative applications with LLMs, offering a rich perspective for research on LLMs based on knowledge graphs.

\subsubsection{Handling Contradictory Memories}

In long-term conversations, the management and construction of memory is a complex and important task, especially when it comes to conflicting memories. It is commonly believed that avoiding conflicting memories is necessary \cite{commonsense}. However, this view does not entirely conform to the true characteristics of human cognition. Human memory is highly context-dependent. The same person may recall different information in different contexts, and even memories of the same event may vary with changes in context. Therefore, allowing memories with conflicting interpretations to coexist within an individual's memory system not only aligns with the natural laws of human cognition but also adds richness and authenticity to conversations.
To effectively deal with these conflicting memories, the author has proposed several optimization strategies for large-language models to choose from flexibly in long-term conversations. First is the “Memory Resolution” strategy. Inspired by entity resolution methods, this strategy merges conflicting memories into an information-rich sentence based on the context of the memory sources, thereby cleverly resolving these conflicts. Second is the “Memory Disambiguation” strategy. The contradiction between two statements may arise from the absence of context, that is, pragmatic ambiguity. Drawing on entity disambiguation methods, the memory disambiguation strategy clarifies the true meaning of each memory by extracting relevant information from the context of the memory, thereby eliminating ambiguity. Finally, there is the “Retention” strategy. Due to the limitations of natural language inference (NLI) models, sometimes memories predicted to be contradictory may not actually be so, but rather a deviation in the model's understanding. In such cases, retaining these seemingly conflicting memories can provide more possibilities and depth to the conversation.
Through these strategies, the author aims to transform conflicting memories into sentences containing richer information about the speaker, thereby enhancing the quality and coherence of long-term conversations.

\subsection{Memory Utilization}
\begin{figure}[H]
  \begin{center}
    \includegraphics[width=1.0\linewidth]{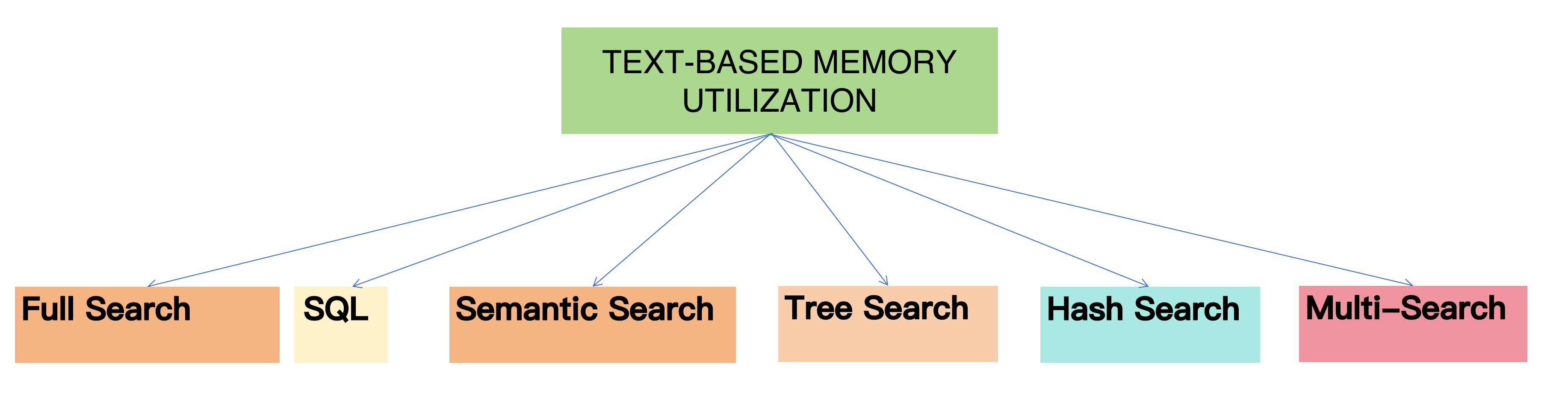}
  \end{center}
  \caption{Text-based memory utilization}
  \label{fig:mem_util}
\end{figure}

The overall structure of Text-based memory utilization is shown in Figure \ref{fig:mem_util}. The classification in the figure is based on the search method. Each method is introduced below.

\subsubsection{Full-text Search}

Full-text search \cite{comedy} involves scanning the entire text dataset to locate specific character sequences or strings. This method not only looks for exact matches but also considers approximate matches based on text structure and wording. In MAS, when facing broad queries or when the exact location or label of information is unknown, the agent can use a full-text search. For example, when a user asks a broad question about “the impact of climate change,” the agent can utilize a full-text search to browse its entire memory and extract relevant paragraphs or interactions.

\subsubsection{Structured Query Language}

Structured Query Language (SQL) is also a method for search \cite{comedy}.
Tabular databases can be queried using standardized specialized programming languages, such as SQL or data frame-based Python languages (like pandas). Recently, a highly efficient method for querying tabular data has outperformed existing text2SQL methods. This algorithm creates a small library of functions for manipulating and retrieving elements in a table. By using a large language model (LLM), these function chains can be called sequentially to perform advanced multi-hop queries on the table. The authors applied the linked list structure to metadata queries of conversational logs. They used two functions to retrieve subsets of rows from the table:
 \texttt{f\_value(column\_name, [value1, value2,...])}
 \texttt{f\_between(column\_name, [value1, value2])}
The \texttt{f\_value} function retrieves all rows where at least one of the listed values matches in the specified column. The \texttt{f\_between} function retrieves all rows where the values are between \texttt{value1} and \texttt{value2}. Although these functions are simple, the authors found that they can accurately complete all questions in the test set in principle. The authors adapted the prompts from the original linked list method to this custom function library. They used the LLM to write function chains, with separate prompts to write 1) the function name, 2) the first parameter (column name), and 3) the second parameter (value).

\subsubsection{Semantic Search}
\cite{comedy} also proposed semantic search.
Given the compressed memory \( \hat{M} \) and the incoming dialogue \( D_t \), GPT4-Turbo outputs a memory-based response. Annotators then review and optimize these responses, focusing on aspects such as relevance, coherence, and personalization. They ensure that each annotated response \( c_{t+1} \) accurately reflects the user's current state and previous interactions, maintaining high memorability and engagement.

\subsubsection{Tree-based Search}
The Markov Decision Process (MDP) is defined by the tuple $(S, A, P, R, \gamma)$, where $S$ represents the set of tree nodes, and $A$ is the aggregation function that generates parent node text from child node text. The goal of $A$ is to create concise summaries of the key information from child nodes for meaningful Hierarchical Abstract Trees (HATs). Whenever a new child node is added, $A$ updates the tree to maintain consistency. In this implementation, GPT is used as the aggregation function to summarize characters from child node text, with specific prompts detailed in the appendix.

GPT is chosen for its conditional text generation ability, which helps find the optimal traversal path in the HAT based on node text and user queries. The framework is open and general, allowing the memory agent to be a neural network, an RL agent, or a GPT approximation. The study uses a multi-session chat dataset with conversations between two speakers on a messaging platform. Sessions are short but resume after pauses to continue or start new topics. The number of episodes per session is listed in a table.

When a node in the HAT is updated, $\sigma$ updates the parent node, propagating changes upward. The memory agent's task is to find the best traversal path in the HAT based on the user query $q$, with GPT generating optimal actions for traversal.

\subsubsection{Hash-based Search}
\cite{thinkinmemory} proposed hash-based search methods.
In the retrieval task based on memory storage, in order to obtain the most relevant thoughts, the author has designed a two-stage retrieval paradigm. When faced with a new query \( Q \), the first step is to obtain the embedding vector \( x \) of the query using a large language model agent. This model is capable of mapping the query text into a high-dimensional vector space to capture its semantic information, providing a foundation for the subsequent retrieval process.
Subsequently, the retrieval process based on Locality-Sensitive Hashing (LSH) commences. LSH is an efficient approximate nearest neighbor search technique that uses specific hash functions to map similar data points into nearby hash buckets, thereby reducing the number of data points that need to be compared. The design of the hash functions is the core of LSH, and different methods may employ different function designs, such as those based on random projection. Typically, multiple hash functions are used to enhance the accuracy and robustness of the retrieval.

During the retrieval, the hash value of the query vector \( x \) is calculated, and the hash buckets that are the same or similar to it are searched for. The data points in these buckets are the candidate's similar data points. This method can quickly locate the content semantically similar to the query without the need to compare the entire dataset one by one. However, LSH provides approximate results, which may have some errors. But by reasonably designing the hash functions and adjusting parameters, such as the number of hash tables and the combination of hash functions, the errors can be reduced to a certain extent while maintaining the efficiency of the retrieval.
The LSH function is as follows, 
\begin{equation}
 \text{hash}(x) = \text{LSH}(x) \label{eq:lsh} 
 \end{equation}
the hash index of the query can be generated, thus laying the foundation for the subsequent retrieval operations.

\subsubsection{Multiple Searches}
In the process of retrieval of memories, conducting multiple searches is indeed a good strategy. Through repeated searches, the search scope can be gradually narrowed down, and the accuracy of retrieval can be improved, thereby better meeting the user's needs. For example, in some complex query scenarios, the initial search may only find part of the relevant information. However, by conducting multiple searches and combining the results of the previous search, the query conditions can be further optimized to obtain more comprehensive and accurate information. In addition, multiple searches can also help language models better understand the user's needs, thereby generating answers that are more in line with user expectations.

\cite{self-reflection} employs a dense retrieval method, using the SimCSE model to encode both user input and each memory record into vectors. By calculating the cosine similarity, the K-relevant top records are selected for response generation. However, since the retrieved memory records may not all be useful and could even be completely irrelevant, the author introduces a self-reflection mechanism for memory. Specifically, after obtaining the retrieval results, the language model is first required to determine whether this information is sufficient to respond to the user's input. If it is sufficient, the language model extracts the relevant parts from the memory records as output. If not, the original query needs to be improved by generating keywords or phrases that represent the missing information, which are then expanded into the original query for another round of retrieval. This self-reflection process can be repeated multiple times until the language model finds sufficient information.

\section{KV Cache-Based Memory}

This chapter introduces long-term memory based on the KV cache. Since transformer-based LLMs are currently the mainstream, especially with the dominance of the decoder-only structure, KV cache-based memory has become a hot research topic, particularly in the field of long context. In terms of the structure of this chapter, we follow the structure of Chapter 3, introducing memory from three aspects: acquisition, management, and utilization. Since there are many research works related to memory acquisition, we divide it into two sections: KV selection and KV compression. KV selection refers to choosing some KVs to save and discarding the rest. We categorize the selection strategies into eight types: regularity-based summarization, score-based methods, special token embeddings, learning-based KV selection, different selection strategies for different layers and heads, identical KV cache for different layers and heads, locality-sensitive hashing strategy, and backtracking-based KV selection. For KV compression in memory acquisition, we summarize it as low-rank compression, KV merging, multimodal compression, KV quantization, and multi-level compression.

For the second major part, memory management, we divide it into offload strategies (that is, placing memory on cheaper memory), optimization strategies combined with the OS system, storage data structures, and shared attention. We also introduce an article that explores Hash-based Search. 

For the utilization of memory, that is, memory retrieval, we introduce efficient attention, locality-sensitive hashing, vector retrieval methods, and other optimization methods.
\begin{figure}[H]
  \begin{center}
    \includegraphics[width=1.0\linewidth]{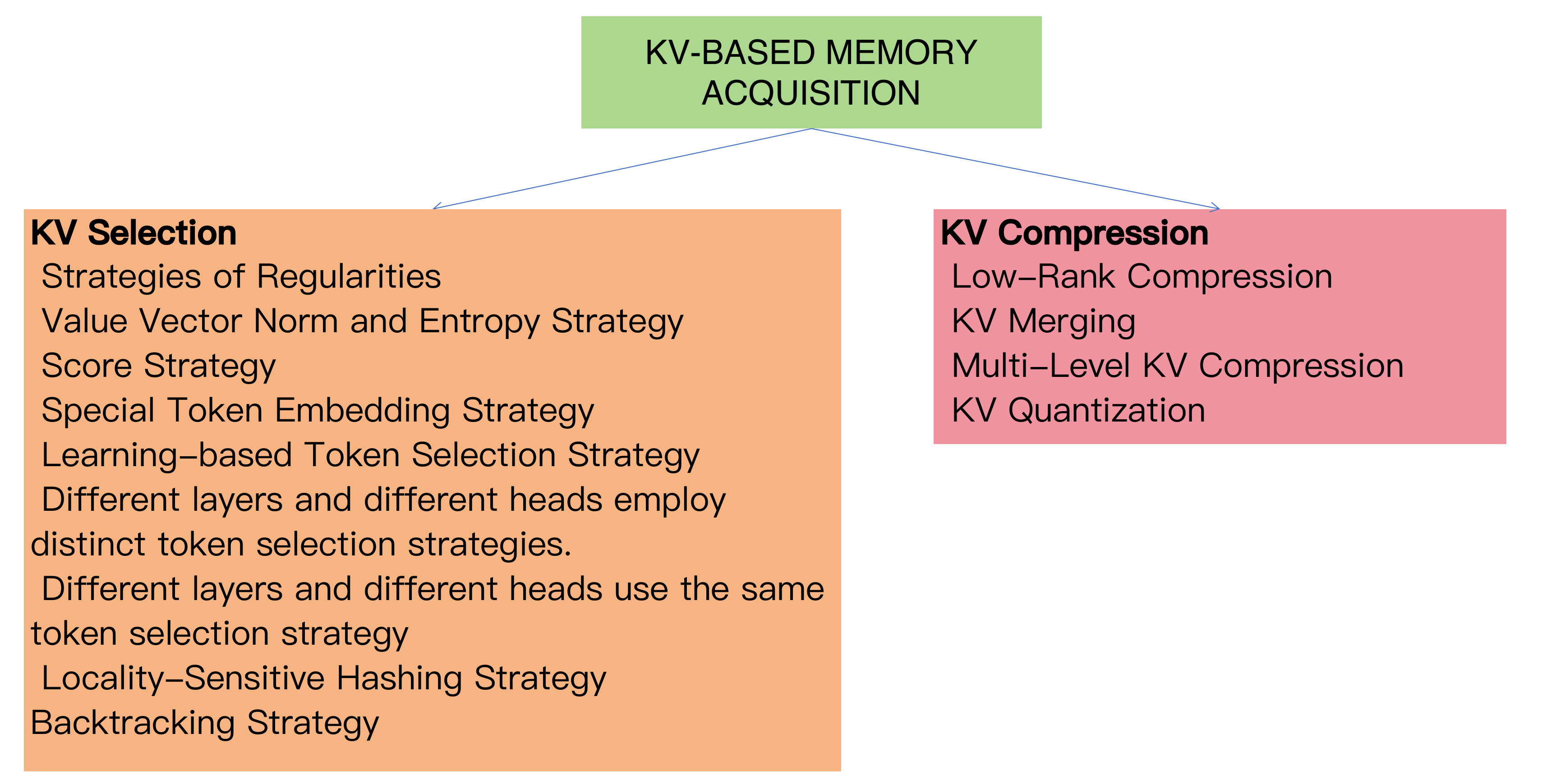}
  \end{center}
  \caption{KV-cache based memory management}
  \label{fig:kv_acquisition}
\end{figure}

As shown in Figure \ref{fig:kv_acquisition}, KV acquisition can be divided into two parts: KV selection and KV compression. 
KV selection means selecting the most valuable token and evicting other tokens.
KV selection means selecting the most valuable token and expelling other tokens.

\subsection{Memory Acquisition: KV Selection}
\subsubsection{Strategy of Regularity}
Traditional KV cache eviction strategies prioritize the eviction of less critical KV pairs based on attention scores. However, these strategies often degrade the quality of generation, leading to issues such as context loss or hallucinations. To address this problem, \cite{d2o} introduce Dynamic Discriminative Operations (D2O), a novel method that optimizes the size of KV caches without fine-tuning while retaining key context. First, by observing the different densities of attention weights between shallow and deep layers, the authors leverage this insight to determine which layers should avoid excessive eviction to minimize information loss. Subsequently, for the eviction strategy of each layer, D2O innovatively introduces a compensation mechanism that maintains a similarity threshold to re-evaluate the importance of previously discarded tokens and decide whether they should be recalled and merged.

\cite{longt5} introduces a text-to-text transformer called LongT5, which is capable of efficiently processing long sequences of text. LongT5 incorporates two new attention mechanisms: Local Attention and Transient Global Attention (TGlobal). These mechanisms enhance the model's ability to handle long sequences while retaining the original characteristics of T5. In particular, the TGlobal attention mechanism, by introducing relative position bias and layer normalization parameters, significantly improves the model's performance in long-sequence tasks.

\cite{scissorhands} speculate that not all tokens need to be stored in memory for large language models to understand context. Just as humans can skim an article and grasp its main ideas, large language models may also be able to skim and understand. It is commonly observed that the attention scores of a token follow a strong power-law distribution, meaning that a token only pays high attention to a few other tokens. More importantly, the authors have observed repeated attention patterns among different tokens in a sequence in trained large language models. Some tokens are more important throughout a paragraph. Specifically, for two different tokens, there is a similarity in the objects they highly attend to, as well as in the objects they ignore. Inspired by these observations, the authors propose the hypothesis of importance persistence: Only those tokens that have had a significant impact in the previous step will have a notable impact in future steps. If this hypothesis holds true, it implies that it is possible to anticipate which tokens may be important for future generations.

\cite{sequence} proposes a KV cache eviction strategy called CORM (Cache Optimization with Recent Message). CORM can dynamically retain key-value pairs important for the inference process without the need for model fine-tuning. This method optimizes KV caching and token generation for LLMs by leveraging the attention information of the most recent query.
The core idea of the CORM method is that if the recent query vectors are sufficiently similar, the current query can directly utilize the attention information of the most recent query. Moreover, if certain key-value pairs appear to be less informative for the most recent query, removing these pairs during the generation process can significantly preserve the model's performance.

\cite{efficientstr} first discovered an interesting phenomenon known as the Attention Sink, where autoregressive LLMs tend to allocate a large number of attention scores to the initial tokens, even if these tokens are not semantically important. Based on this phenomenon, StreamingLLM retains a small number of KV pairs for the initial tokens as attention sinks and combines them with sliding-window KVs to anchor the attention computation and stabilize model performance.

\cite{longnet} introduce LONGNET, a variant of the Transformer that can scale sequence lengths to over 1 billion tokens without sacrificing performance on shorter sequences. Specifically, the authors propose a dilated attention mechanism, which exponentially expands the attention range as the distance increases. LONG NET has several significant advantages: 1) It has linear computational complexity, with logarithmic dependencies between any two tokens in the sequence; 2) It can serve as a distributed trainer for extremely long sequences; 3) Its dilated attention mechanism is a direct replacement for standard attention and can be seamlessly integrated into existing Transformer-based optimization processes.

\cite{longformer} added "global attention" at pre-selected input positions. Importantly, the authors made this attention operation symmetric: that is, tokens with global attention will attend to all tokens in the sequence, and all tokens in the sequence will also attend to it.

ETC (Extended Transformer Construction) \cite{etc} is a novel Transformer architecture designed to address two key challenges of standard Transformer architectures when dealing with long sequences and structured inputs. ETC introduces a global-local attention mechanism to scale attention to longer inputs. In this architecture, the input is divided into two sequences: the global input and the long input. The global input typically contains a smaller number of auxiliary tokens, while the long input includes the standard input sequence that a Transformer would normally process. 
The global-local attention mechanism restricts tokens in the long input to attend only to the global input or their local neighbors, thereby reducing computational complexity. This approach changes the attention computation from quadratic to linear scaling.

\cite{empower} primarily introduces a technique called Attention Transition, which enables the understanding of longer contexts by adjusting attention weights. Through experiments, the authors discovered that when information is passed between different layers of the model, many identical tokens appear, and their positions gradually move closer together. This indicates that important information is progressively transmitted to nearer positions during the layer computation process. Additionally, the authors found that rotary embeddings limit the understanding of long-range information, but crucial information tends to move closer together during the layer computation process.

\subsubsection{Value Vector Norm and Entropy Strategy}
Traditional methods typically rely solely on attention scores to determine the importance of key tokens. However, \cite{attentionscore} finds that the value vectors of key tokens also influence their importance. Therefore, the authors propose a novel token pruning method called Value-Aware Token Pruning (VATP), which takes into account both the attention scores and the norms of the value vectors, providing a more comprehensive metric for assessing token importance.
When considering how to evaluate and prune tokens to reduce the memory footprint of the KV cache, the authors not only focus on the attention scores but also pay particular attention to the norms of the value vectors. Specifically, they introduce a new method called “Value-Aware Token Pruning (VATP),” which combines the attention scores and the norms of the value vectors to assess the importance of each token.

\subsubsection{Score Strategy}
Previous studies have proposed KV cache compression techniques that identify unimportant tokens based on accumulated attention scores and remove them from the KV cache, noting that only a small number of tokens play a significant role in the attention operation. However, \cite{a2sf} observed that existing accumulated attention scores are not suitable for the Transformer decoder architecture. In decoder models, the number of times attention scores are accumulated varies with the order in which tokens appear due to the masking effect, leading to an uneven comparison among different tokens.
To address this issue, the authors propose a technique called Accumulated Attention Scores with Forgetting Factor (A2SF), which introduces a "forgetting factor" in the process of accumulating attention scores. A2SF repeatedly multiplies the past attention scores generated by older tokens by the forgetting factor, thereby imposing a penalty on past attention scores over time. Consequently, older tokens receive a greater penalty, providing fairness among tokens of different ages. By fairly comparing tokens, the authors can more effectively select important tokens.

One important feature of the Quest algorithm \cite{quest} is its ability to dynamically adjust the contents of the KV cache. Specifically, as the query vector Q changes, Quest re-estimates the criticality of tokens and updates the contents of the KV cache accordingly. This dynamic adjustment ensures that the KV cache always contains the key information required for the current query.
The authors' insight is that to avoid missing critical tokens, they should select pages that contain tokens with the highest attention weights. However, to efficiently select pages, they should compute an approximate attention score based on this insight. The authors found that the upper bound of attention weights within a page can be used to approximate the highest attention within that page. The upper bound of attention weights can be calculated using the channel-wise minimum (mi) and maximum (Mi) of the key vectors. Given a query vector Q, Quest computes the maximum possible value for channel i by taking Ui = max(Qimi, QiMi). Note that Ui is always greater than the product of Qi and any key value Ki on that page, regardless of the sign of Qi. Therefore, when the authors sum Ui, they obtain the upper bound of attention weights for all key vectors on that page.
After deriving the upper bound of attention weights, the authors select the top K pages as key pages, where K is an arbitrarily defined hyperparameter. To demonstrate the feasibility of Quest, the authors performed actual self-attention calculations and collected the Top-K attention scores for each page. As shown in Figure 3, the authors' query-aware sparsity mostly aligns with the predicted sparsity. Quest only performs normal self-attention on the selected pages, which significantly reduces memory movement. The authors define the number of tokens in the selected pages as the "token budget."

Adaptive Token Release \cite{efficientsparse} is a mechanism that evaluates token importance to select which tokens to release. First, the model leverages the Top-K attention mechanism to calculate attention weights and identify the top K important tokens, thereby reducing computational load while maintaining performance comparable to that of full-attention models. When updating the cache's key-value (KV) states, the model adds the latest KV states to the cache and selectively removes an older, less important KV state to maintain a stable cache size. Additionally, the model employs a dynamic eviction strategy based on the current context and task requirements to flexibly choose which tokens to release, ensuring that the token release strategy can adaptively adjust to different tasks and contexts.

Keyformer \cite{keyformer} leverages the observation that approximately 90\% of attention weights in generative inference are concentrated on a specific subset of tokens, referred to as "key" tokens. Keyformer identifies these key tokens using a novel scoring function and retains only the key tokens in the KV cache.

The scoring function of Keyformer is as follows:
\begin{equation}
f_{\theta}(x_i) = \frac{e^{x_i + \zeta_i / \tau}}{\sum_{j=1}^{k} e^{x_j + \zeta_j / \tau}}
\end{equation}
Where \(x_i\) represents the unnormalized logits, \(\zeta_i\) is the added Gumbel noise distribution, and \(\tau\) is the temperature parameter used to adjust the smoothness of the probability distribution. This scoring function, proposed in Keyformer, integrates Gumbel noise distribution into the unnormalized logits to address issues caused by token dropping.

Sparse Window Attention (SWA) \cite{alisa} is a method used in the inference process of large language models (LLMs), aiming to reduce memory usage by generating sparse attention patterns while maintaining model accuracy. The core of SWA lies in combining local static and global dynamic sparse patterns: it generates static patterns on local tokens, retaining the most recent tokens to maintain the sequential semantics of the language while capturing the dynamically changing semantic importance of previous tokens through dynamic patterns. This hybrid pattern can more effectively capture key tokens in the sequence.

In \cite{h2o}, the authors propose a novel implementation of KV caching that significantly reduces its memory footprint. Their approach is based on a striking observation: a small subset of tokens contributes the majority of the values when computing attention scores. The authors refer to these tokens as "Heavy Hitters" (H2). Through a comprehensive investigation, the authors find that (i) the emergence of H2 is natural and closely related to the frequent co-occurrence of tokens in the text, and (ii) removing them leads to a significant drop in performance. Based on these insights, the authors introduce "Heavy Hitters Oracle" (H2O), a KV cache eviction strategy that dynamically maintains a balance between recent and H2 tokens.

\cite{sirllm} use large language models to calculate the entropy measure for each input token, thereby assessing its importance. Subsequently, tokens with higher entropy values, which are considered key tokens, are retained and stored in the KV cache. This method enhances the model's memory capacity in infinitely long stream conversations.

\subsubsection{Special Token Embedding Strategy}
Previous studies have attempted to alleviate this issue by selectively discarding tokens, but \cite{razorattention} irreversibly deletes critical information that may be required for future queries. In this paper, we propose a novel KV cache compression technique that retains all token information. Our research shows that: i) most attention heads primarily focus on local context; ii) only a small number of attention heads, referred to as retrieval heads, can substantially attend to all input tokens. These key observations prompt us to adopt separate caching strategies for attention heads. Consequently, we introduce RazorAttention, a training-free KV cache compression algorithm that preserves full caches for these critical retrieval heads and discards remote tokens in non-retrieval heads. Additionally, we introduce a new mechanism involving "compensation tokens" to further recover information from the discarded tokens.

\cite{landmark} proposed in this paper maintains the random access flexibility of the Transformer by dividing long texts into consecutive chunks and using attention mechanisms to retrieve relevant chunks. By introducing special marker tokens to represent each chunk, the degree of attention to their corresponding chunks can be controlled through the attention mechanism.

In \cite{taking}, the authors propose a simple yet effective method that enables large language models to "take a deep breath," encouraging them to summarize information from discrete text chunks. Specifically, the authors divide the text into multiple chunks and insert a special token <SR> at the end of each chunk. They then modify the attention mask to integrate the information of each chunk into the corresponding <SR> token. This allows large language models to not only interpret information from individual historical tokens but also to interpret information from the <SR> tokens, thereby aggregating the semantic information of the chunks.

\subsubsection{Learning-based Token Selection Strategy}
\cite{learnedtoken} mainly introduces a new learning-based token pruning method called Learned Token Pruning (LTP), which can adaptively remove unimportant tokens from the input sequence and learn a threshold for each layer within the Transformer layers. The LTP method can adaptively change the length of the pruned sequence. LTP employs a learning-based thresholding approach for pruning. First, for each token, it calculates the average attention probability across all attention heads and defines this average as the importance score of the token. Then, a learnable threshold is set for each layer, and if a token's importance score is below the threshold for that layer, the token will be pruned in that layer. To make the threshold learnable, LTP adopts a soft pruning scheme, using a Sigmoid function to create a differentiable soft mask. Although this soft mask approximates the behavior of hard pruning, it allows gradients to flow to the threshold parameters, enabling the learning and optimization of the thresholds.

\cite{sparse} introduces an efficient sparse attention mechanism called SPARSEK Attention, designed to address computational and memory efficiency issues in long-range Transformer computations. This mechanism achieves linear training complexity and constant inference-time memory cost by selecting a fixed number of key-value pairs for each query. SPARSEK Attention incorporates learnable sparse patterns, learning sparsity in a data-driven manner rather than using fixed patterns. This enables SPARSEK Attention to better adapt to different tasks and data distributions.

SparseK implements sparsity through a differentiable top-k selection operation, which allows the selection of key KV pairs while maintaining gradient optimization. The core idea of SparseK is to relax the sparsity constraint from a simple probability simplex to a k-sum constraint, defined as follows:
\begin{equation}
\text{SparseK}(\mathbf{z}, k) \coloneqq \operatorname{arg\,min}_{\mathbf{p} \in \mathbb{C}} \|\mathbf{p} - \mathbf{z}\|^2
\end{equation}
where \(\mathbb{C}\) is the k-sum constraint set, defined as:
\begin{equation}
\mathbb{C} = \left\{ \mathbf{p} \mid \mathbf{0} \leq \mathbf{p} \leq \mathbf{1}, \mathbf{1}^\top \mathbf{p} = k \right\}
\end{equation}
Moreover, the solution of SparseK can be expressed as:
\begin{equation}
\mathbf{p}^* = \max\left(\min\left(\mathbf{z} - \tau(\mathbf{z}), \mathbf{1}\right), \mathbf{0}\right)
\end{equation}
Here, \(\tau(\mathbf{z})\) is a threshold function that satisfies
\begin{equation}
\sum \mathbf{p}^* = k
\end{equation}
and \(\mathbf{z}\) represents the sorted coordinates. This solution is achieved through a soft thresholding operation, allowing the model to perform actual selection operations in the forward pass while maintaining gradient updates in the backward pass.

\cite{dynamic} can dynamically prune the context in Transformer models to enhance computational efficiency and interpretability. The authors introduce a parameterized pruning mechanism at each layer, enabling the model to selectively discard context information that is no longer needed.
The specific pruning process can be expressed with the following formulas:

For each pair of tokens \(i\) and \(j\) in the sequence, calculate a pruning gate control signal \(\overline{I}_{n,j}^{\ell}\):
   \begin{equation}
   \overline{I}_{n,j}^{\ell} = \sigma \left( \frac{(\mathbf{Q}_{\text{int}}^{\ell})_n^\top (\mathbf{K}_{\text{int}}^{\ell})_j}{\sqrt{r}} + \beta^{\ell} \right)
   \end{equation}
   Here, \(\sigma\) is the sparse sigmoid function, \(\mathbf{Q}_{\text{int}}^{\ell}\) and \(\mathbf{K}_{\text{int}}^{\ell}\) represent the query and key vectors of the interaction, respectively, \(\beta^{\ell}\) is the learnable bias parameter, and \(r\) is the scaling factor.
Use the gate control signal \(\overline{I}_{n,j}^{\ell}\) to determine whether to retain the influence of token \(j\) in the context on the generation of token \(n\):
   \begin{equation}
   I_{k,j}^{\ell} = \prod_{n=j+1}^{k} \overline{I}_{n,j}^{\ell}
   \end{equation}
   If \(j < k\), otherwise \(I_{k,j}^{\ell} = 1\) if \(j = k\), or \(I_{k,j}^{\ell} = 0\) if \(j > k\).
Through this method, the model can dynamically adjust its context when generating each new token, thereby reducing unnecessary computation and memory usage.

\subsubsection{Different layers and different heads employ distinct token selection strategies.}
\cite{squeezeattention} mainly introduces an algorithm called SQUEEZEATTENTION, which is used to manage KV caches in LLM (Large Language Model) inference to further reduce the memory consumption of inference. The paper points out that existing KV cache optimization methods mostly focus on optimizing from the sequence dimension or the batch dimension while neglecting the potential opportunities in the attention layer dimension. By analyzing the behavior of different attention layers during inference, the authors propose a method for precise re-allocation of cache budgets for attention layers, thereby further reducing the total amount of KV caches.

\cite{think:} focuses on long-context scenarios and addresses the inefficiency of KV cache memory consumption during inference. Unlike existing methods that optimize memory based on sequence length, the authors find significant redundancy in the channel dimension of KV caches, characterized by imbalanced magnitude distributions and low-rank structures in attention weights. Based on these observations, the authors propose ThinK, a novel query-dependent KV cache pruning method that aims to minimize attention weight loss while selectively pruning the least important channels.

\cite{moa} propose a Mixture of Attention (MoA), which automatically customizes different sparse attention configurations for various heads and layers. MoA constructs and navigates a search space of various attention patterns and their scaling rules relative to input sequence lengths. It profiles the model, evaluates potential configurations, and identifies the optimal sparse attention compression plan. MoA adapts to different input sizes, revealing that some attention heads expand their focus to accommodate longer sequences, while other heads consistently concentrate on fixed-length local contexts.

\cite{zebra} propose a novel model architecture called Zebra. This architecture effectively addresses the quadratic time and memory complexity issues caused by full attention in Transformers by employing grouped local-global attention layers. The authors' model, resembling the alternating stripes of a zebra, balances local and global attention layers, significantly reducing computational demands and memory consumption.

LONGHEADS \cite{longheads} has been proposed to enhance the ability of pre-trained language models (LLMs) to handle long contexts without additional training. The core idea of the paper is to fully exploit the potential of multi-head attention by restricting each attention head to select and focus on important context blocks within the pre-trained length, thereby avoiding the processing of tokens that exceed the pre-trained length (OOD problem). Additionally, the paper proposes a block selection strategy that leverages the model's inherent dot-product attention to select important blocks for each head.

SnapKV \cite{snapkv} is a KV cache compression method for optimizing large language models (LLMs) when processing long contexts. It is implemented through the following steps: First, an observation window located at the end of the prompt is used to monitor the model's attention patterns on input tokens during the generation process, which are stable and consistent throughout the generation. Next, a voting algorithm based on the observation window is employed to identify important positions that are consistently attended to by attention heads during generation, thereby selecting the KV pairs corresponding to the most important tokens. Before voting, SnapKV introduces a pooling-based clustering technique to capture complete local information and filter out irrelevant tokens, thus compressing the context while preserving its integrity. Finally, the selected KV pairs are concatenated with the observation window to form a new, smaller KV cache. The model only needs to perform attention calculations on this significantly reduced cache during subsequent generations, thereby improving efficiency.

In \cite{pyramidkv}, the authors investigate whether the attention-based information flow within large language models aggregates information for long contexts through significant patterns. The authors' observations reveal that large language models aggregate information in a pyramidal fashion, where attention is widely dispersed at lower layers, gradually integrates within specific contexts, and eventually focuses on key tokens (i.e., tokens with high activation values or attention sinks) at higher layers. Based on these insights, the authors developed PyramidKV, a novel and efficient KV cache compression method. This method dynamically adjusts the size of KV caches at different layers, allocating more cache at lower layers and less at higher layers, in contrast to traditional methods that maintain a consistent KV cache size.

Upon re-examining the current pruning methods, \cite{optimizing} found that these methods essentially minimize the upper bound of the L1 pruning loss between the outputs of the multi-head self-attention mechanism before and after pruning. Moreover, their analysis revealed that the common practice of evenly distributing the budget across attention heads compromises the generation quality after pruning. Based on these findings, the authors propose a simple yet effective adaptive budget allocation algorithm. This algorithm not only optimizes the theoretical loss upper bound but also reduces the L1 pruning loss in practice by matching the different characteristics of each head.

\subsubsection{Different layers and different heads use the same token selection strategy}

DeepSeek-V2 \cite{deepseek} employs an innovative architecture, including Multi-head Latent Attention (MLA) and DeepSeekMoE. MLA achieves efficient inference by significantly compressing the Key-Value (KV) cache into latent vectors, while DeepSeekMoE enables the training of powerful models at an economical cost through sparse computation.

\cite{effectively} explored the low-rank characteristics of the KV cache and proposed a new method for compressing KV heads. In particular, the author carefully optimized the transition from MHA to GQA to minimize compression error. Additionally, to maintain compatibility with Rotary Position Embeddings (RoPE), the author introduced a specialized strategy for the key cache with RoPE.

In \cite{simple}, the authors propose a simple and effective KV cache compression strategy based on the L2 norm. They first examined the relationship between the L2 norm of the cached keys and the attention scores, finding a strong correlation between them. Specifically, key embeddings with low L2 norms tend to produce high attention scores during the decoding process. Based on this observation, the authors propose a KV cache compression strategy that retains only the keys with the lowest L2 norms and their corresponding values. A significant advantage of this strategy is that it does not require additional training or substantial modifications to the model, and can be directly applied to any Transformer-based decoder model.

The main content of \cite{crosslayer} introduces a method named LISA (Learnable Sharing Attention), which aims to reduce the redundant attention computations across layers in large language models. The authors demonstrate through experiments that in most cases, the attention weights between different layers exhibit similarity, and thus the efficiency of the model can be improved by sharing these weights. To achieve this goal, the authors propose two key components: the Attention Head Alignment Module and the Difference Compensation Module.

\cite{mlkv} primarily introduces a layer-compressed KV cache method for efficient inference in large language models. The method significantly reduces the memory consumption of KV caches and improves inference speed by decreasing the number of cached layers.

\subsubsection{Locality-Sensitive Hashing Strategy}

Reformer \cite{reformer} is a paper co-authored by researchers from Google Research and the University of California, Berkeley, focusing on improving the efficiency of Transformer models when dealing with long sequence data. The paper proposes two key technologies: First, Locality-Sensitive Hashing (LSH), which replaces the traditional dot-product attention mechanism. By randomly projecting keys into multiple "buckets" and computing attention only between keys within the same bucket, LSH reduces the complexity of self-attention from O(L²) to O(LlogL), significantly decreasing the computational load while maintaining similarity. Second, Reversible Residual Layers, which alter the forward propagation of the network so that activation values need to be stored only once during training, rather than at each layer as in traditional Transformers, thereby greatly reducing memory usage. Additionally, Reformer introduces Chunking, which splits the activations in the feed-forward network into multiple chunks for separate processing, further reducing memory consumption.

\cite{faster} primarily introduces a type of attention mechanism called Sparse Flash Attention, which improves computational efficiency when processing large sequences. The authors optimize the attention matrix by introducing a dynamic sparse structure and hash algorithms, thereby reducing the amount of computation.

\subsubsection{Backtracking Strategy}
Previous token pruning methods typically remove tokens during the forward propagation phase without considering their impact on the attention of subsequent layers, which may lead to the loss of tokens that are important for a given task. To address this issue, \cite{sparse} proposes an attention backtracking method that tracks the importance of each attention in the Transformer architecture from the output to the input, in order to retain tokens that have a significant impact on the final prediction.

\subsection{Memory Acquisition: KV Compression}

\subsubsection{Low-Rank Compression}
KV-Cache compression methods typically sample a subset of effective tokens or quantize the data into lower numerical bit widths. However, these methods cannot exploit the redundancy in the hidden dimension of the KV tensors. \cite{palu} presents a hidden dimension compression approach called Palu, a novel KV-Cache compression framework that leverages low-rank projection. Palu decomposes linear layers into low-rank matrices, caches the compressed intermediate states, and dynamically reconstructs the full keys and values. To improve accuracy, compression rate, and efficiency, Palu further includes a medium-grained low-rank decomposition scheme, an efficient rank search algorithm, low-rank-aware quantization compatibility enhancements, and optimized GPU kernels with operator fusion.

\cite{getmore} is an efficient technique for optimizing KV caches in large language models (LLMs). It achieves cache compression by combining sparse strategies with low-rank kernels. Specifically, LESS first employs a sparse strategy to selectively cache a subset of KV pairs based on certain criteria (such as importance or frequency), thereby reducing the size of the cache. Subsequently, it learns the residuals between the original attention outputs and the approximated outputs from the sparse strategy and accumulates this residual information into a fixed-size low-rank cache. This low-rank cache does not scale with the sequence length, thus effectively reducing memory usage. In this way, LESS not only retains key information but also allows queries to access information discarded by the sparse strategy. As a result, it minimizes performance degradation while maintaining a small cache size. Additionally, LESS can synthesize multiple tokens during generation to produce longer and more coherent sequences. Moreover, both its attention computation and cache update processes are carefully designed.

\subsubsection{KV Merging}
In \cite{modeltell}, the authors propose a novel KV cache merging method called KVMerger, which aims to achieve adaptive KV cache compression for long-context tasks without significantly degrading performance under a limited memory budget. The authors’ method is inspired by an interesting observation: within a single sequence, key states exhibit high similarity at the token level. To facilitate merging, the authors develop an efficient and straightforward merging set identification algorithm to recognize KV states that are suitable for merging. The authors’ merging set identification algorithm gives rise to a second observation: from the perspective of similarity, the sparsity of KV caches is dataset-agnostic and persists at the model level. Subsequently, the authors propose a Gaussian kernel-weighted merging algorithm to selectively merge all states within each merging set.
The formalized expression of KV cache merging is as follows:
\begin{enumerate}
    \item \textbf{Identification of Mergeable Sets}: Formulate the KV cache merging problem into a constrained clustering issue and use an effective mergeable set identification method to recognize suitable KV states for merging.
    \item \textbf{Weighted Merging Algorithm}: For each identified mergeable set, use the Gaussian kernel weighted merging algorithm to merge states. The expressions for the merged key state ($M(\mathcal{K})$) and value state ($M(\mathcal{V})$) are as follows:
    \begin{equation}
    M(\mathcal{K}) = \bigcup_{i=1}^{d} F(\mathcal{S}_{ki})
\end{equation}
\begin{equation}
    M(\mathcal{V}) = \bigcup_{i=1}^{d} F(\mathcal{S}_{vi})
\end{equation}
    where $\mathcal{S}_{ki}$ and $\mathcal{S}_{vi}$ represent the sub-mergeable sets of key states and value states, respectively, and $F$ is the merging function that maps each mergeable set to a single state.
    \item \textbf{Gaussian Kernel Weighted Merging}: Within each identified mergeable set, use the Gaussian kernel as a weight to merge states, ensuring that the merged information is retained without significantly reducing the generative performance of the LLM.
\end{enumerate}

\cite{transformerfam} focuses on introducing the concept of working memory based on attention mechanisms into deep learning and explores methods for implementing working memory in Transformer models. By incorporating feedback attention mechanisms, Transformer models are able to pass prior information to the current module, thereby enabling effective processing and information compression of long text contexts.

\cite{efficientcontent} introduces an efficient content-based sparse attention mechanism called Routing Transformer. Traditional attention mechanisms face high computational complexity when dealing with long sequences. In contrast, the Routing Transformer addresses this issue by introducing sparsity and clustering techniques, thereby maintaining the model's flexibility while improving computational efficiency.

\subsubsection{Multi-Level KV Compression}
\cite{hmt} propose the Hierarchical Memory Transformer (HMT), a novel framework that enhances the model's ability to handle long contexts by emulating human memory behavior. With memory-augmented segment-level recurrence, the authors construct a memory hierarchy by retaining tokens from early input segments, passing memory embeddings along the sequence, and recalling relevant information from history.
Initially, the long input text sequence is divided into multiple segments \( X_i \), each containing a fixed number of tokens. For the first segment \( X_0 \), the model encodes it using a Transformer encoder to obtain the initial memory:
\begin{equation}
\text{Mem}_0 = \text{TransformerSegment}(X_0)
\end{equation}
This initial memory serves as the foundation for processing subsequent segments.
For each subsequent segment \( X_i \), the model updates the memory in a hierarchical manner. Specifically, the model first encodes the current segment \( X_i \) to obtain the local memory:
\begin{equation}
\text{LocalMem}_i = \text{TransformerSegment}(X_i)
\end{equation}
Then, the model combines the local memory with the previous memory state and selects important information through an attention mechanism to update the global memory:
\begin{equation}
\text{GlobalMem}_i = \text{Attention}(\text{LocalMem}_i, \text{Mem}_{i-1})
\end{equation}
To control the size of the memory and retain key information, the model compresses the global memory using a compression module:
\begin{equation}
\text{Mem}_i = \text{Compression}(\text{GlobalMem}_i)
\end{equation}
This hierarchical memory update mechanism allows the model to dynamically update and compress the memory while processing each new segment, thereby avoiding information overload.
When processing each new segment, the model not only generates the encoding of the current segment but also recalls relevant information from the memory and combines it with the output of the current segment:
\begin{equation}
\text{Output}_i = \text{TransformerSegment}(X_i) + \text{Attention}(X_i, \text{Mem}_i)
\end{equation}
This recall mechanism enables the model to enhance the representation of the current segment by leveraging historical information, thereby better understanding the contextual relationships in long texts.
After processing all segments, the model concatenates the outputs of each segment to form the final representation of the entire long text:
\begin{equation}
\text{Final Output} = \text{Concat}(\text{Output}_0, \text{Output}_1, \dots, \text{Output}_n)
\end{equation}
where \( n \) is the number of segments in the input sequence.


\subsection{Memory Management}

\begin{figure}[H]
  \begin{center}
    \includegraphics[width=1.0\linewidth]{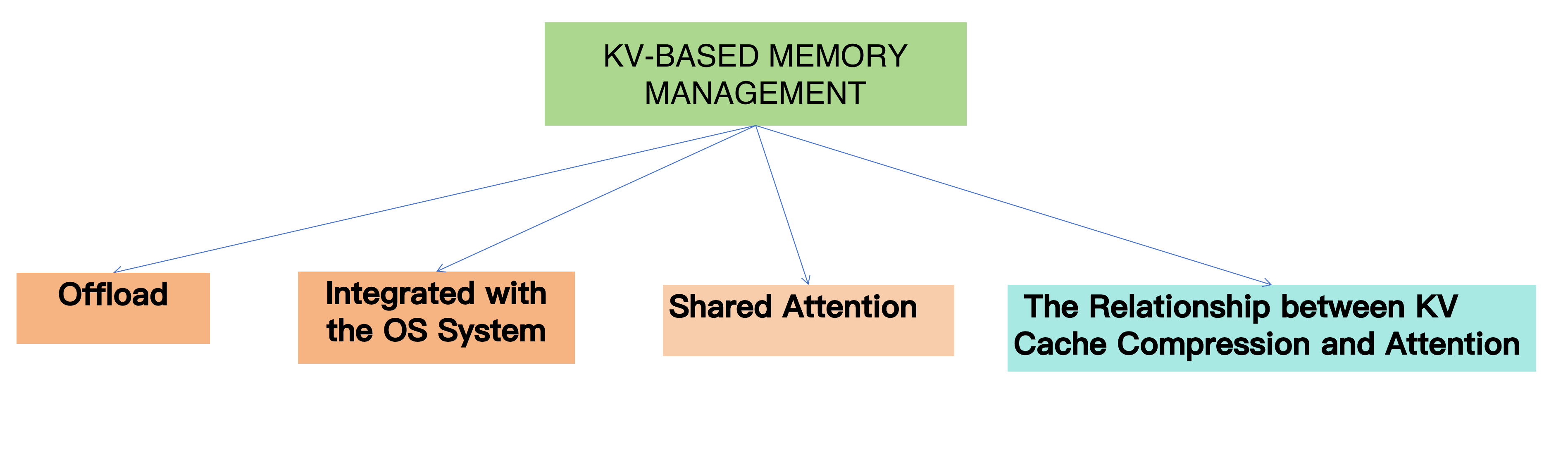}
  \end{center}
  \caption{KV-cache based memory management}
  \label{fig:kv_management}
\end{figure}
KV-cache-based memory management can be divided into four parts: offload, integrated with the OS system, shared attention, and the relation between KV cache compression and attention, as shown in Figure \ref{fig:kv_management}.

\subsubsection{Offload}

Attention offload for LLMs is an optimization strategy that improves inference efficiency and reduces costs by offloading attention computations to specialized memory-optimized hardware. LLM inference consists of a prefilling stage and a generation stage, with the latter demanding high memory bandwidth and suffering from reduced accelerator utilization as context length increases and KV caches expand. Researchers propose leveraging inexpensive memory-optimized devices to handle attention operations, while using high-end accelerators for other parts, creating a heterogeneous setup to boost performance and cost efficiency. Experiments show that the Lamina system, which employs this strategy, achieves 1.48 to 12.1 times higher throughput per dollar compared to traditional solutions. Additionally, techniques such as GPUDirect RDMA, device-side busy polling, and communication-attention overlap strategies have been employed to reduce scheduling and network latency, thereby optimizing user experience. Overall, attention offload technology holds great promise for improving resource utilization efficiency and reducing costs.

\cite{instinfer} introduces a novel long-context low-power machine learning inference system called InstInfer. This system improves inference performance while reducing energy consumption and costs by offloading the crucial attention computation tasks during the decoding phase from GPUs to CSDs (Computational Storage Devices), by executing attention computations within in-storage.

\cite{dynamic-mem} focuses on improving existing large language models (LLMs) through Dynamic Memory Compression (DMC) to accelerate inference speed. Researchers introduce decision variables and importance variables into the model, enabling it to learn customized compression strategies, thereby reducing memory usage during the inference process.

The batch size is limited by some intermediate results that are repeatedly used, namely the KV-Cache. These intermediate results occupy too much memory space, preventing more sequences from being processed simultaneously on the graphics processing unit (GPU). Although it is possible to offload them to host memory, the CPU-GPU bandwidth is an inevitable bottleneck. \cite{fastdecode} have discovered a method to decompose the Transformer model into two parts with different characteristics, one of which includes memory-bound KV-Cache access. The key insight of the authors is that the aggregated memory capacity, bandwidth, and computational power of CPUs across multiple nodes is an efficient choice for processing this part. The performance improvement comes from reduced data transfer overhead and increased throughput of the GPU in processing the other part of the model.

\subsubsection{Memory Management with OS System}

\cite{memserve} propose MemServe, a unified system that integrates both inter-request and intra-request optimizations. MemServe introduces MemPool, an elastic memory pool for managing distributed memory and KV caches across service instances. By utilizing the MemPool API, MemServe is the first to combine context caching with fragmented inference, supported by a global scheduler that enhances cache reuse through a locality-aware policy based on a global prompt tree.

\cite{vattention} introduces a dynamic memory management system called vAttention for LLM (Language Model) service systems. attention leverages the operating system's support for virtual memory and demand paging to reduce the programming burden and performance loss associated with traditional PagedAttention methods.

 High-throughput serving of large language models requires processing a sufficient number of requests in a single batch. However, existing systems perform poorly in this regard because the memory footprint of the key-value cache (KV cache) for each request is substantial and dynamically grows and shrinks. If not managed properly, this memory can be severely wasted due to fragmentation and redundant replication, thereby limiting the batch size. To address this issue, \cite{efficientmem} propose PagedAttention, an attention algorithm inspired by the classic virtual memory and paging techniques in operating systems. Building on this, the authors develop vLLM, a large language model serving system that achieves (1) near-zero waste of KV cache memory and (2) flexible sharing of KV caches both within and across requests to further reduce memory usage.

\cite{magicdec} introduces a technique called Speculative Decoding, which balances the trade-off between latency and throughput by performing speculative decoding during the generation of long contexts. Through theoretical modeling and empirical analysis, the authors find that Speculative Decoding can increase throughput, reduce latency, and maintain the accuracy of generation when processing longer sequences and large batch requests.

\cite{infinite} introduces Infinite-LLM, a novel LLM (Large Language Model) serving system designed to address the highly dynamic context length management in LLM requests. The paper first analyzes the computational characteristics of LLM models, pointing out the limitations of traditional static model parallelization and KVCache scheduling methods when dealing with dynamic context lengths. To tackle these issues, the paper proposes three innovations: the DistAttention mechanism, which distributes attention computation and KVCache across the entire GPU cluster to improve cluster throughput; the liability mechanism, which allows borrowing memory from other instances to handle large context tasks and increases generation throughput; and gManager, which is used for global planning and coordination of request and KVCache allocation.

\cite{efficientllm} mainly introduces an efficient inference technique called KCache for the inference process of large language models (LLMs). KCache stores the K Cache in high-bandwidth memory (HBM) and the V Cache in CPU memory during inference. It dynamically selects which key information to copy back from CPU memory to HBM for computation based on the softmax results of the attention calculation. This approach ensures the accuracy of inference while improving its performance.

\subsubsection{Shared Attention}
Large language models typically precompute KV caches to speed up prefilling when processing multi-text-chunk inputs, but \cite{cacheblend} is only effective when the text chunk is a prefix of the input. When the text chunk is not a prefix, the precomputed KV cache cannot be directly used due to the lack of cross-attention with the preceding text, limiting cache reuse.
To address this issue, the authors propose a scheme called CacheBlend. CacheBlend reuses precomputed KV caches regardless of whether the text chunk is a prefix or not. It selectively recomputes the KV values of a small subset of tokens to partially update the reused KV caches and supplement the cross-attention. The delay for recomputing these tokens can be pipelined with the retrieval of KV caches, allowing CacheBlend to store caches in slower, higher-capacity devices without increasing inference delay. This approach maintains the same generation quality as full prefill while improving efficiency.

\cite{beyond} mainly introduces a method called Shared Attention (SA), which reduces the computational and storage overhead in large language models by sharing attention weights. Traditional methods have primarily focused on sharing key-value caches to reduce memory overhead, but still require independent computation of attention weights in each layer. In contrast, the SA method directly shares the computed attention weights, which not only significantly reduces the size of key-value caches but also decreases the computational load during model inference.

\subsubsection{Relationship between KV Cache Compression and Attention}

The attention mechanism is a core component of the Transformer architecture and has enabled the remarkable success of large-scale language models. However, the theoretical underpinnings of the attention mechanism are not yet fully understood, especially its non-convex optimization dynamics. In \cite{max}, the authors investigate the pioneering softmax attention model \( f(X) = \langle Xv, \text{softmax}(XW p) \rangle \), where \( X \) is a sequence of tokens and \( (v, W, p) \) are training parameters. They prove that running gradient descent on \( p \), or equivalently on \( W \), leads to directional convergence towards a maximum-margin solution that can distinguish between locally optimal tokens and non-optimal tokens. This explicitly formalizes attention as an optimal token selection mechanism. In particular, the authors' results apply to general data and precisely characterize the optimality of tokens in terms of the value embeddings \( Xv \) and the geometry of the problem. The authors also provide a broader analysis of the regularization path, establishing the maximum-margin property of attention even in the presence of a non-linear prediction head. When \( v \) and \( p \) are optimized simultaneously and logistic loss is used, the authors determine the conditions under which the regularization path converges in the direction of their respective hard-margin support vector machine solutions, where \( v \) separates the input features based on the labels.


\section{Parameters-Based Memory}
\begin{figure}[H]
  \begin{center}
    \includegraphics[width=1.0\linewidth]{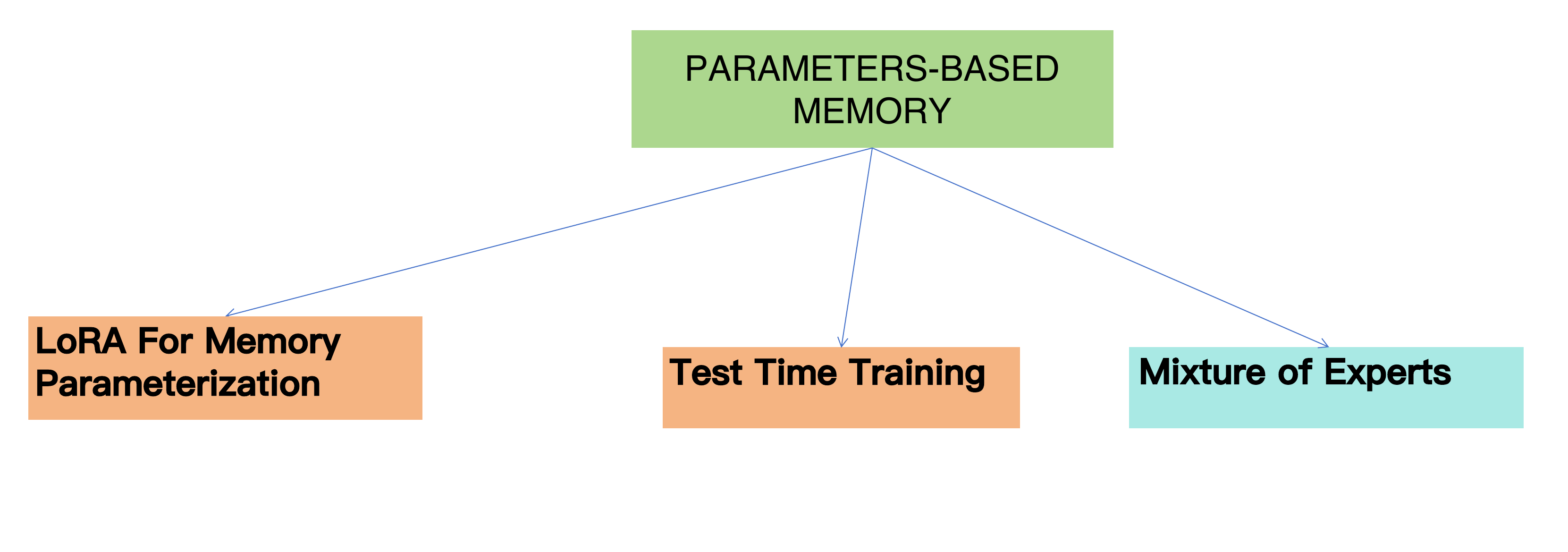}
  \end{center}
  \caption{Parameter based memory}
  \label{fig:parameter_based_memory}
\end{figure}
Both text-based memory and KV cache-based memory are merely storage of historical information, and they do not become the model's true "memory." In this section, we introduce a method of memory parameterization, which transforms previous memories into parameters as part of the model itself.

We categorize the strategies of memory parameterization into three types: LoRA, Test Time Training (TTT), and Mixture of Experts (MoE), and introduce each of them respectively, as shown in Figure \ref{fig:parameter_based_memory}.

\subsection{LoRA For Memory Parameterization}
In the online scenario, Transformer language models face the challenge of expanding context. As the length of the context increases, the attention mechanism requires more and more memory and computational resources, which significantly reduces the throughput of the language model. To address this issue, \cite{compressed} proposes a context compression memory system. This system continuously compresses the accumulated attention key/value pairs into a compact memory space, enabling the language model to perform efficient inference within the limited memory of the computing environment. The author's compression process integrates lightweight conditional LoRA (Low-Rank Adaptation) into the forward pass of the language model during inference, without the need to fine-tune the entire weight set of the model. In addition, the author models the recursive compression process as a single parallelized forward computation, thereby achieving efficient training. Through evaluations on tasks such as dialogue, personalization, and multitask learning, the author demonstrates that this method can achieve the performance level of a full-context model while reducing the context memory size by a factor of 5. Furthermore, the author shows the applicability of this method in streaming scenarios with infinite context length, where its performance is superior to traditional sliding window methods.

The author assumes a Transformer language model \( f_{\theta} \) with \( L \) layers, where the dimension of the hidden state is \( d \). To simplify the notation, the author sets the length of the compressed token to 1. It is worth noting that the length of the compressed token can be extended to any length. Under these assumptions, the total size of the attention keys/values for the compressed token (denoted as \( \langle \text{COMP} \rangle \)) is \( 2 \times L \times d \). At each time step \( t \), the author appends the \( \langle \text{COMP} \rangle \) token to the context \( c(t) \) and performs attention operations on the keys/values of \( c(t) \) and the previous memory state \( \text{Mem}(t-1) \) using the \( \langle \text{COMP} \rangle \) token. Through this operation, using the attention keys/values of the \( \langle \text{COMP} \rangle \) token, the author obtains the compressed hidden features \( h(t) \in \mathbb{R}^{2 \times L \times d} \).

\subsection{Test Time Training For Memory Parameterization}
Test Time Training (TTT) \cite{test-1,test-2,test-3} is a technique that optimizes model performance through additional training steps during the testing phase. It is primarily used to address situations where the test distribution deviates from the training distribution. The core idea of TTT is to fine-tune the model using information from the test data itself during the testing phase, thereby better adapting to the test data distribution. Specifically, TTT introduces an auxiliary task and updates certain model parameters, typically those of the feature extractor, by optimizing the loss function of the auxiliary task.

During the training phase, the model learns not only the main task but also an auxiliary task simultaneously. For example, in an image classification task, the auxiliary task could be predicting the rotation angle of an image. The main task and the auxiliary task share a portion of the model's parameters, usually a feature extractor. During the testing phase, for each test sample, the model fine-tunes the shared feature extractor by optimizing the loss function of the auxiliary task, updates the model parameters, and then proceeds with the prediction for the main task.

The formulas for TTT can be expressed as follows: Suppose the model parameters are denoted by $\theta$, the loss function for the main task by $l_m$, and the loss function for the auxiliary task by $l_s$. Let the parameters of the shared feature extractor be $\theta_e$, the parameters specific to the main task be $\theta_m$, and the parameters specific to the auxiliary task be $\theta_s$. During the training phase, the model's optimization objective is to minimize the weighted sum of the losses from the main and auxiliary tasks:
\begin{equation}
    \min_{\theta_e, \theta_m, \theta_s} \alpha l_m(\theta_e, \theta_m) + (1 - \alpha) l_s(\theta_e, \theta_s)
\end{equation}
where $\alpha$ is the weight that balances the losses of the two tasks. During the testing phase, for each test sample $x$, the model updates the feature extractor parameters by optimizing the loss function of the auxiliary task:
\begin{equation}
    \theta_e' = \theta_e - \eta \nabla_{\theta_e} l_s(\theta_e, \theta_s; x)
\end{equation}
where $\eta$ is the learning rate.

TTT has several advantages. It can dynamically adapt to the test data distribution by adjusting the model parameters according to the test data, thereby improving the model's adaptability to the current test sample. Additionally, TTT has low computational overhead since it only performs a small amount of training on a small number of samples, making it more efficient. Moreover, TTT is typically unsupervised, relying only on the unsupervised signals from the test samples (such as the loss of the auxiliary task) without requiring additional labeling. TTT is widely applied in fields such as image classification and graph neural networks, especially when there is a deviation between the test and training data distributions, significantly enhancing the model's performance. Through TTT, the model can update its parameters during inference to adapt to the inference environment, without the need for new memory content.

\subsection{MoE For Memory Parameterization}

\cite{branch} introduces an innovative model training method called Branch-Train-MiX (BTX), which aims to efficiently integrate multiple expert large language models (LLMs) into a single mixture-of-experts (MoE) model. This method combines the strengths of the Branch-Train-Merge (BTM) approach and the MoE architecture while mitigating their respective drawbacks.

The BTX method consists of three main steps. First, during the Branch and Train phase, multiple copies (referred to as expert models) are created from a pre-trained seed model and trained independently on different data subsets, each corresponding to a specific knowledge domain such as mathematics, programming, or Wikipedia. This training process is parallel and asynchronous, reducing communication costs and increasing training throughput.
Next, in the MiX phase, the feedforward sublayers of these expert models are merged into a single MoE module to form a unified MoE model. Within each Transformer layer, a router network is used to select which expert's feedforward sublayer should be applied to each token. The weights of the self-attention sublayers and other modules are combined through simple averaging.
Finally, in the MoE Finetuning phase, the merged model is further fine-tuned on the entire training dataset, allowing the router network to learn how to route tokens dynamically between different experts during testing.

The advantages of the BTX method include the parallel and asynchronous nature of the expert training phase, which reduces communication costs and enhances training efficiency. The final BTX model is a unified neural network that can be fine-tuned like any standard LLM. Additionally, the model's FLOPs (floating-point operations) during inference do not increase significantly because it is sparsely activated, despite the increase in parameter count. The paper also explores various variants of BTX, such as load balancing, different routing methods, and strategies for splitting and merging experts, to further improve model performance and efficiency.

\section{Hidden-State-Based Memory}
\begin{figure}[H]
  \begin{center}
    \includegraphics[width=1.0\linewidth]{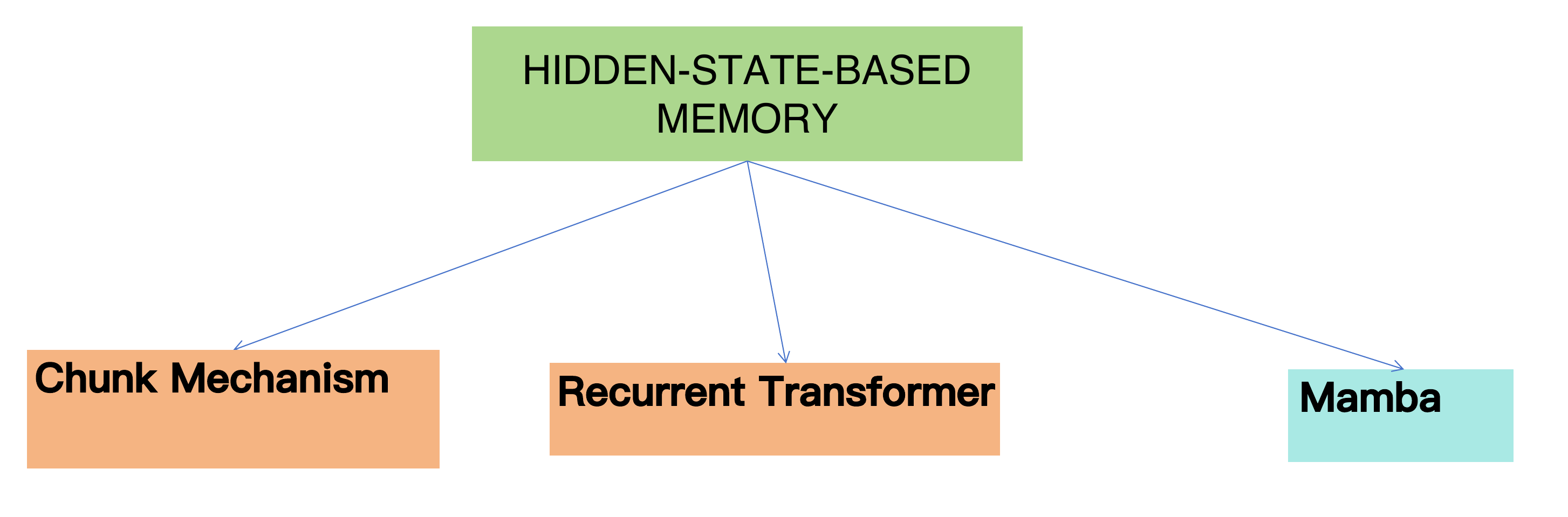}
  \end{center}
  \caption{Hidden state-based memory}
  \label{fig:hidden}
\end{figure}
In addition to retaining text memory, KV cache, and parametrized memory, some researchers have combined the idea of hidden states in RNNs with current methods. These methods can be categorized into three types: methods based on the chunk mechanism, recurrent transformer methods, and methods based on Mamba. Figure \ref{fig:hidden} shows the overall structure.

\subsection{Chunk Mechanism}
When processing long texts, LLMs employ a method known as the chunk mechanism, which involves dividing the long text into smaller chunks to enable more efficient processing and embedding. This process begins by segmenting the text into multiple chunks based on fixed character counts, sentences, paragraphs, or semantics. It ensures that tokens with coherent semantics are aggregated together, while tokens with different semantics are separated to maintain the semantic integrity of each chunk. Subsequently, each chunk is individually fed into the language model for processing, and the model generates an embedding vector for each chunk, which captures the semantic information of the chunk. To avoid losing contextual information at the boundaries of chunks, an overlapping strategy is often used, meaning that the beginning part of each chunk overlaps with the ending part of the previous chunk. These embedding vectors of the chunks can be utilized for downstream tasks such as information retrieval and content recommendation.


The context window in Transformers serves as active memory for tasks like few-shot learning and conditional generation, which rely heavily on prior context tokens. However, increasing context length leads to quadratic growth in computational cost. Recent methods combining fixed-size sliding windows with initial tokens achieve linear complexity but unconditionally evict all tokens from the KV cache at the window's end, causing information loss. To address this, a new mechanism \cite{training} is proposed: maintaining cascaded sub-buffers to store longer context while keeping the same total cache size, with each buffer conditionally accepting important tokens evicted from the previous one.

The previous process leads to a fixed heuristic pattern of discarding tokens, which is not ideal as the model may naively retain less valuable tokens while discarding important ones. To solve this, the authors dynamically select important tokens to retain by tracking their average attention scores using exponential moving averages (EMA). They discard tokens with lower EMA scores. Specifically, given a hyperparameter $\gamma \in [0, 1]$ and attention score vector $\mathbf{s}_k^{(t)}$ at time step $t$, the average attention score $\mu^{(t)}$ is updated as:
\begin{equation}
\mu^{(t+1)} = \gamma \mu^{(t)} + (1 - \gamma) \mathbf{s}_k^{(t)}.
\end{equation}

The main content of \cite{citrus} introduces a method called CItruS, which aims to address the issue of information loss in long-sequence modeling. The authors propose two sub-processes: the language modeling process and the task-solving process. In the language modeling process, a chunk-based state eviction method is employed to enhance modeling efficiency. In the task-solving process, instruction-aware state eviction is introduced, utilizing the hidden state of the final instruction as an additional instruction-aware query to extract and retain task-relevant information. Task-specific responses are generated using this key-value cache.

\cite{finch} proposes a novel method called FINCH, which compresses input context by leveraging the weights of a pre-trained self-attention model. Given a prompt and a long text, FINCH iteratively identifies key (K) and value (V) pairs relevant to the prompt, based on chunks of the text. Only these pairs are stored in the KV cache, which is space-limited within the context window and ultimately contains a compressed version of the long text.

\cite{unlimiformer} introduces a method called Unlimiformer, which can accept input texts of arbitrary length during testing. Unlimiformer constructs the hidden states of the input text using a k - k-nearest neighbors index. It then queries the k - nearest neighbors index in the standard cross-attention heads of each decoder layer, using the k - nearest neighbor distances as attention scores and focusing only on the top - k input tokens. In this way, Unlimiformer is able to extend existing encoder-decoder Transformer models to accept inputs of arbitrary length. This method not only has low computational and memory overhead but also accurately approximates global attention.

\cite{adapting} introduces a method called AutoCompressor, which extends the context window of language models by generating summary vectors, thereby enhancing the model's ability to handle long texts. The summary vectors are generated from long documents and are used to improve language modeling for subsequent passages. The authors demonstrate through experiments that summary vectors can encode useful information, aid in downstream task performance, and reduce the inference cost of context learning. Additionally, the authors showcase the effectiveness of using precomputed summary vectors for text retrieval and reranking tasks on large corpora.

\cite{megabyte} introduces a multiscale decoder architecture called MEGABYTE, which can perform end-to-end differentiable modeling of sequences exceeding one million bytes. MEGABYTE divides the sequence into patches and employs local submodels within the patches and a global model between the patches. This approach enables subquadratic self-attention, larger feed-forward layers, and improved parallelism during decoding, thereby achieving better performance at a lower cost during both training and generation.

\cite{chunk} introduces a simple method for handling long sequences that can be applied in Transformer models. Traditional Transformer models face limitations in computational complexity when dealing with long sequences. However, the method proposed in this paper can effectively handle long sequences while maintaining good performance on short sequences. Specifically, the paper proposes a method of dividing long inputs into manageable-length blocks and selecting the most representative tokens for decoding. In addition, the paper also introduces a selector based on policy learning to optimize the compression process of long sequence information.

\cite{chunkattention} mainly introduces an attention module called ChunkAttention, which employs prefix-based KV caching and a two-stage partitioning method to improve the efficiency of self-attention. Specifically, the paper proposes using a prefix tree to implement KV caching, which can dynamically detect and remove redundant KV caches to reduce memory usage. Meanwhile, the paper also describes a two-stage self-attention computation method, consisting of chunk-first and sequence-first stages.

\cite{training-free} introduces a novel training-free framework called Dual Chunk Attention (DCA) for extending the context window of large language models (LLMs). The paper first describes the three components of DCA: intra-chunk attention, inter-chunk attention, and successive-chunk attention. These attention mechanisms help the model effectively capture both long-range and short-range dependencies when processing long sequences.

In \cite{efficientlong}, the authors propose SLED: Sliding Encoder and Decoder, a simple method for handling long sequences that leverages battle-tested pre-trained LMs for short texts. Specifically, the authors divide the input into overlapping chunks, encode each chunk using a short-text LM encoder, and fuse information between chunks using a pre-trained decoder (fusion in the decoder).

\subsection{Recurrent Transformer}
The Recurrent Transformer is a hybrid model that combines the strengths of Recurrent Neural Networks (RNNs), such as their ability to retain information when processing long sequences, with the parallel processing and global dependency-capturing capabilities of the Transformer model. This type of model typically includes an embedding layer that maps input words to high-dimensional vectors, a recurrent layer such as LSTM or GRU to capture long-term dependencies, and self-attention mechanisms to process all elements of the sequence in parallel. Additionally, feed-forward networks perform nonlinear transformations on the outputs of the attention layers, while layer normalization helps stabilize the training process. Residual connections are formed between the outputs and inputs of each sublayer to avoid the vanishing gradient problem. The Recurrent Transformer may include encoder and decoder components, and some models adopt hierarchical structures to handle sequences at different levels. In the model, information flows through the recurrent layer to capture long-term dependencies, then through the attention mechanism to understand the global relationships between elements, and finally through the feed-forward network for further transformation. Different variants of the Recurrent Transformer may adjust these components or introduce new mechanisms according to specific applications and research objectives to enhance performance.


RWKV \cite{rwkv} is a novel neural network architecture designed specifically for natural language processing (NLP) tasks, combining the strengths of recurrent neural networks (RNNs) and Transformer architectures. It is uniquely structured to efficiently process long text sequences while balancing long-term context memory and parallel computing capabilities.

The core structure of RWKV includes input embedding, time gate, state update, and output layer. The input embedding maps each word or character in a text sequence to a vector. The time gate controls the flow of information in the temporal dimension through the following formula:
\begin{equation}
    \text{Time Gate}(x_t) = \sigma(W_t x_t + b_t),
\end{equation}
where \(x_t\) is the input vector at time step \(t\), \(W_t\) is the weight matrix of the time gate, \(b_t\) is the bias vector of the time gate, and \(\sigma\) is the activation function, typically the Sigmoid function.

The state update combines the output of the time gate with the previous time step's state, using the following formula:
\begin{equation}
    h_t = \text{Time Gate}(x_t) \cdot h_{t-1} + (1 - \text{Time Gate}(x_t)) \cdot f(x_t),
\end{equation}
where \(h_t\) is the hidden state at time step \(t\), \(h_{t-1}\) is the hidden state at the previous time step, and \(f(x_t)\) is a nonlinear transformation of the current input, typically implemented as a neural network layer.

Finally, the output layer converts the hidden state into output results through the following formula:
\begin{equation}
    y_t = W_y h_t + b_y,
\end{equation}
where \(y_t\) is the output at time step \(t\), \(W_y\) is the weight matrix of the output layer, and \(b_y\) is the bias vector of the output layer.

The main advantages of RWKV are that it efficiently processes long text sequences, avoiding the vanishing or exploding gradient problems often encountered in traditional RNNs. It also possesses certain parallel computing capabilities, which enhance training and inference efficiency. By dynamically controlling the flow of information, RWKV can better capture long-term dependencies in context.

\subsection{Mamba}
Mamba is a novel sequence model based on the Selective State Space Model (SSM), aiming to combine the advantages of Recurrent Neural Networks (RNNs) and Transformers while improving efficiency and performance. Its core structure consists of three parts: Transformer MLP, SSM Block, and Selective Mechanism. The Transformer MLP is responsible for channel mixing, similar to the feed-forward network part in Transformers; the SSM Block is in charge of sequence modeling, capturing long-range dependencies in sequences through state space models; and the Selective Mechanism dynamically adjusts model parameters based on the input, similar to attention mechanisms but more efficient.

The workflow of Mamba mainly includes the following key steps. First, the input sequence is expanded in dimension through linear projection and then discretized. The discretization formulas are:
\begin{equation}
\bar{A} = \exp(\Delta A)
\end{equation}
\begin{equation}
\bar{B} = (\Delta A)^{-1} (\exp(\Delta A) - I) \cdot \Delta B
\end{equation}
where \(\Delta\) is the selectivity factor that controls the degree of state updates. Next, the Selective Mechanism dynamically generates matrices \(B\), \(C\), and \(\Delta\) based on the input, with formulas:
\begin{equation}
s_B(x) = \text{Linear}_N(x)
\end{equation}
\begin{equation}
s_C(x) = \text{Linear}_N(x)
\end{equation}
\begin{equation}
s_\Delta(x) = \text{Linear}_D(x)
\end{equation}
followed by activation through \(\tau_\Delta = \text{softplus}\). State updates and output computations can be implemented in either recursive or convolutional forms. The recursive form formulas are:
\begin{equation}
h_t = \bar{A} h_{t-1} + \bar{B} x_t
\end{equation}
\begin{equation}
y_t = C h_t
\end{equation}
while the convolutional form formula is:
\begin{equation}
y = K \times x
\end{equation}
where \(K\) is the convolution kernel composed of \(\bar{B}\) and \(C\).




\section{Discussion}

In the current classification of memory mechanisms, text-based memory and memory rooted in model parameters are categorized as long-term memory. These forms of memory can endure across multiple session cycles. In contrast, memory based on KV caches and memory based on hidden states are classified as short-term memory. They exist solely within a single session cycle and vanish once the session concludes.

When compared to human memory mechanisms, current Large Language Models (LLMs) still exhibit significant deficiencies in the realm of information retrieval. A key distinction lies in the fact that humans possess an exceptional ability to summarize and generalize information. Humans can distill complex and intricate matters into simple and clear principles. This process not only significantly reduces the “space” required to store information but also greatly enhances the applicability of memory. The generalized principles can be readily activated and applied across a wide array of different situations, demonstrating remarkable flexibility and generalization capabilities.
Another notable difference is that LLMs lack an inherent ability to forget. Whether it is memory based on text, KV caches, or hidden states, the models will retain all previous information intact in the absence of active human intervention. This leads to the continuous accumulation and expansion of information, which may negatively impact the performance and efficiency of the models. For instance, it can increase storage burdens and reduce retrieval speeds.
Humans also possess the ability to integrate and assimilate knowledge, which represents an advanced form of generalization. When new memories are introduced that are related to or in conflict with existing memories, humans can skillfully integrate the new and old memories to form new memories that replace the original ones. This ability allows human memory to dynamically adapt to new knowledge and information, maintaining its accuracy and relevance.

Among the current methods, short-term memory based on KV caches and hidden states is inherently limited, and Their storage and retrieval capabilities are confined to the current session cycle. On the other hand, text-based memory has theoretical upper limits in terms of both storage capacity and retrieval efficiency. These limitations make it challenging to meet the growing demands of information processing. Therefore, memory parameterization has emerged as a promising direction for the future development of LLM cognitive memory, as it closely resembles human memory mechanisms. The goal of memory parameterization is to enable models to acquire abilities such as summarizing information, integrating knowledge, and selectively forgetting details, akin to human cognitive processes, thereby improving the efficiency of information processing and storage.
Significant progress has been made in the field of memory parameterization. Methods such as LoRA, MOE, TTT, and the design of new modules have been explored. Future research should focus on effectively transforming memory into model parameters without compromising the original memory and without significantly increasing the number of model parameters. This challenging yet highly promising area of research holds the potential to drive significant advancements in LLM cognitive memory, ultimately bringing these models closer to human cognitive capabilities and memory functions.

\bibliographystyle{plainnat}  
\bibliography{egbib}
\end{document}